%% file: main.tex
\documentclass[10pt,twocolumn,letterpaper]{article}

\usepackage[pagenumbers]{iccv} 


\definecolor{iccvblue}{rgb}{0.21,0.49,0.74}
\usepackage[pagebackref,breaklinks,colorlinks,allcolors=iccvblue]{hyperref}
\usepackage{multirow}
\usepackage{colortbl,xcolor}
\usepackage{float}

\def\paperID{8944} 
\def\confName{ICCV}
\def\confYear{2025}

\title{Where do Large Vision-Language Models Look at when Answering Questions?}

\author{
    Xiaoying Xing\textsuperscript{1,2,$\dagger$}, 
    Chia-Wen Kuo\textsuperscript{1}, 
    Li Fuxin\textsuperscript{3}, 
    Yulei Niu\textsuperscript{1}, 
    Fan Chen\textsuperscript{1}, \\
    Ming Li\textsuperscript{1}, 
    Ying Wu\textsuperscript{2}, 
    Longyin Wen\textsuperscript{1}, 
    Sijie Zhu\textsuperscript{1,*} \\
    \textsuperscript{1}Bytedance Intelligent Creation, \textsuperscript{2}Northwestern University \textsuperscript{3}Oregon State University
}

\begin{document}
\maketitle
\let\thefootnote\relax\footnotetext{$*$ Corresponding author, sijiezhu@bytedance.com}
\let\thefootnote\relax\footnotetext{$\dagger$ This work was done during the first author’s internship at ByteDance}

\input{0_abstract}  
\input{1_intro}
\input{2_related}
\input{3_method}
\input{4_experiment}
\input{5_conclusion}
{
    \small
    \bibliographystyle{ieeenat_fullname}
    \bibliography{main}
}
\input{6_appendix}

\end{document}

%% file: 0_abstract.tex
\begin{abstract}
Large Vision-Language Models (LVLMs) have shown promising performance in vision-language understanding and reasoning tasks. 
However, their visual understanding behaviors remain underexplored.
A fundamental question arises: to what extent do LVLMs rely on visual input, and which image regions contribute to their responses?
It is non-trivial to interpret the free-form generation of LVLMs due to their complicated visual architecture (\eg multiple encoders and multi-resolution) and variable-length outputs. In this paper, we extend existing heatmap visualization methods (\eg iGOS++~\cite{khorram2021igos++}) to support LVLMs for open-ended visual question answering. We propose a method to select visually relevant tokens that reflect the relevance between generated answers and input image.
Furthermore, we conduct a comprehensive analysis of state-of-the-art LVLMs on benchmarks designed to require visual information to answer. Our findings offer several insights into LVLM behavior, including the relationship between focus region and answer correctness, differences in visual attention across architectures, and the impact of LLM scale on visual understanding.
The code and data will be released at \url{https://github.com/bytedance/LVLM_Interpretation}.
\end{abstract}
\vspace{-0.4cm}

%% file: 1_intro.tex
\section{Introduction}
\label{sec:intro}
 \begin{figure}[!htb]
    \centering
    \includegraphics[width=0.47\textwidth]{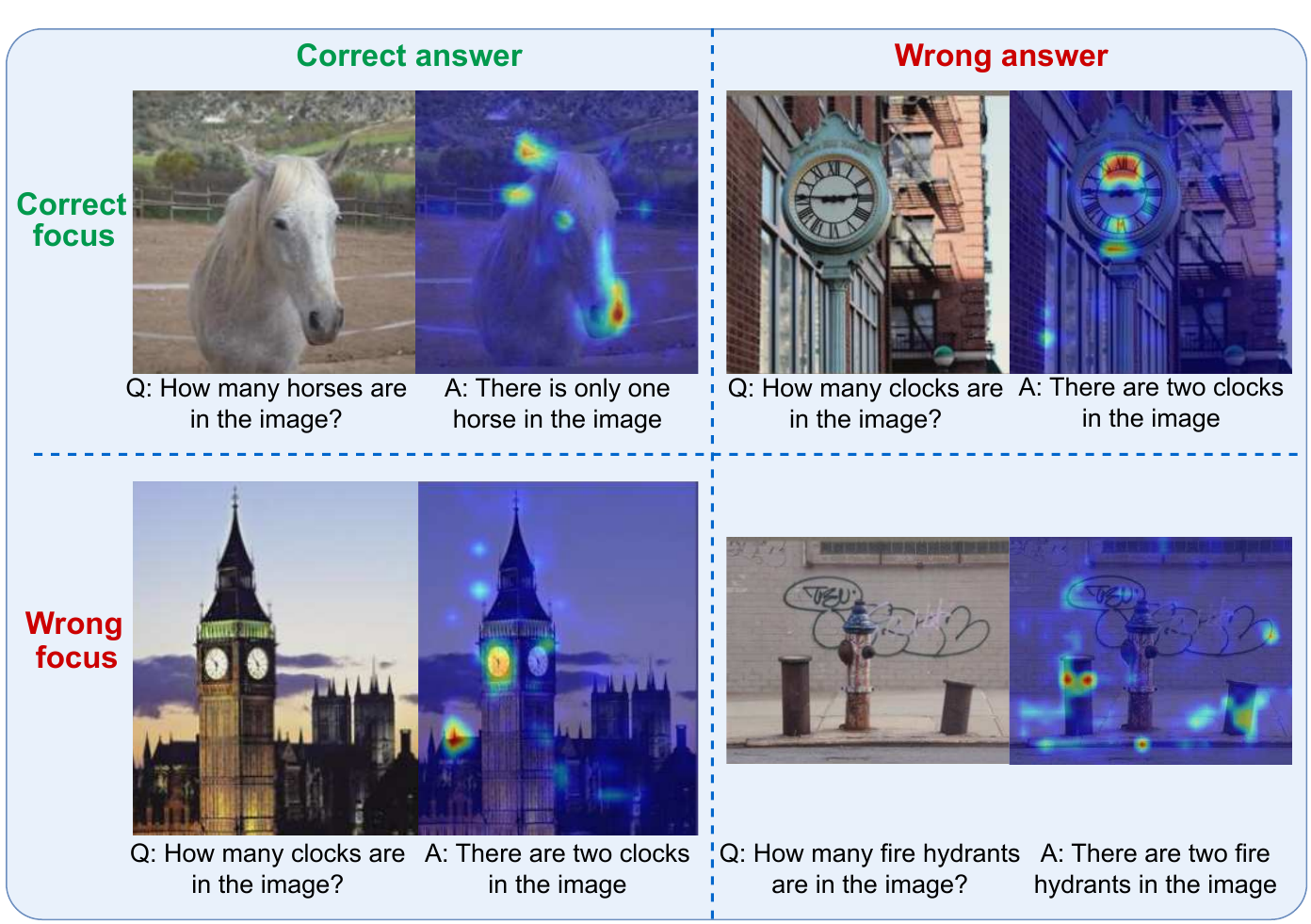}
    \caption{
    Focus regions of LLaVA-1.5 when answering counting questions. The model may correctly focus on the relevant region and produce the correct answer (top left), or it may fail despite attending to the right region due to misinterpretation (top right). In some cases, incorrect focus leads to wrong answers (bottom right), while occasionally, the model answers correctly despite attending to irrelevant regions (bottom left), highlighting challenges in visual grounding and generalization.
    }
    \label{fig:teaser}
\end{figure}

The emerging Large Vision and Language Models (LVLM)~\cite{llavaov,bai2023qwen,team2023gemini,dai2023instructblip} have exhibited strong visual instruction following abilities and achieved remarkable performance on multimodal tasks, 
such as Visual Question Answering (VQA)~\cite{antol2015vqa}. 
Despite different design choices and implementation details, most LVLMs follow the representative visual instruction tuning~\cite{llava} paradigm to align the visual features from pre-trained vison encoders~\cite{clip,dinov2} to a pre-trained LLM~\cite{touvron2023llama,vicuna}.
This enables LVLMs to incorporate visual understanding while retaining the rich knowledge and reasoning abilities of LLMs.

However, the underlying mechanisms behind the visual understanding capabilities of LVLMs remain unclear. Beyond evaluating model performance on various benchmarks, it is crucial to interpret where the LVLM focuses when generating responses, as it can provide insights into why an answer is correct or incorrect and facilitate targeted improvements for multimodal tasks. 
For instance, as shown in Figure~\ref{fig:teaser}, an LVLM may attend to the correct region but still misinterpret the content (\eg, the top-right example), fail to locate the relevant region entirely (\eg, bottom-right), or even produce correct answers based on irrelevant regions (\eg, bottom-left), which can lead to poor generalization.

Before the era of LVLMs, a popular way to interpret visual models is to derive a saliency heatmap of the input image, representing the relevance of the image regions to the output~\cite{gradcam,chefer2021transformer,iia}. Despite the rapid growth of LVLM research, little effort has been made to interpret LVLMs. Existing works~\cite{ben2024lvlm,giulivi2024explaining} often explain single-label outputs or individual tokens within a sentence. 
However, LVLMs generate open-ended responses consisting of multiple tokens with variable lengths, requiring a holistic interpretation of the entire output rather than isolated components.
Interpreting open-ended responses of LVLMs has several challenges, (1) Vision-Language Interaction: LVLMs involve intricate interactions between vision and language modalities, and often exhibit strong bias towards the language priors. Hence it is hard to determine the contribution of each modality to the response. 
(2) Autoregressive Generation: Unlike classification models, LVLMs autoregressively generate free-form text, making it difficult to interpret the model behavior considering the entire output. (3) Model Architecture Complexity: Current LVLMs often use multi-resolution or multi-encoder architectures, making it unreliable to align the features across layers to the corresponding spatial regions in the image.

To address these challenges, we propose a method that enables model interpretation for LVLMs and open-ended responses. It reveals the significant image regions that lead to the generated response, providing insights into the reasoning process. To the best of our knowledge, this is the first heatmap visualization method that applies to any LVLM structure and generate a global interpretation for open-ended responses.
Using this method, we conduct in-depth studies of state-of-the-art LVLMs on benchmarks designed to evaluate the visual understanding, and obtain several insights related to the model behaviors. 
(1) Effect of LLM Scale: Simply increasing the LLM size does not significantly alter the model’s visual attention behavior. (2) Impact of Vision Architecture: Different vision architectures could lead to distinct attention patterns. Multi-resolution models tend to focus on finer details, while multi-encoder models often attend to broader regions. (3) Performance vs. Visual Understanding: Performance and visual understanding behavior are not always aligned; current LVLMs may produce correct answers while attending to irrelevant regions, suggesting that accuracy alone is insufficient to comprehensively reflect the model's capabilities.
The main contributions of this work are as follows:
\begin{itemize}
    \item Proposed to extract the visually relevant tokens from the open-ended responses, which are representative of the vision-related parts of the variable-length responses.
    \item Extended existing visual interpretation methods to interpret LVLMs for free-form text output with several technical improvements.
    \item Conducted in-depth studies of LVLM visual behavior, analyzing how different models attend to image regions when answering visual-related questions.
\end{itemize}

%% file: 2_related.tex
\section{Related Work}
\label{sec:related}
\noindent\textbf{Visual Instruction Tuning.}
Benefiting from the advancement of LLM, current LVLMs often seek to equip LLMs with visual understanding capabilities, which is achieved by visual instruction tuning~\cite{llava}. They utilize a modality connector to align the visual embeddings into prompts that the language models can comprehend~\cite{li2023empowering,li2023blip}. Despite the great progress, existing LVLMs still suffer from hallucination~\cite{jiang2024hallucination,liu2023mitigating,zhang2023language,zhou2023analyzing} due to insufficient visual understanding capabilities. Improving the visual capabilities of LVLMs is crucial to their further improvement and application, and several benchmarks~\cite{cambrian,mmvp,mmstar} have been proposed to highlight the growing attention to this issue.

\noindent\textbf{Visual Attention Visualization.}
We mainly focus on a popular family of interpretation methods, which aims to highlight the image regions most relevant to the model's output. Existing heatmap visualization methods can be mainly categorized to gradient-based approaches~\cite{gradcam,zeiler2014visualizing,zhang2018top,simonyan2013deep} and perturbation-based approaches~\cite{dabkowski2017real,khorram2021igos++,qi2019visualizing,fong2017interpretable}. Gradient-based methods often backpropagate the output of the model and generate the saliency heatmap using the gradients or some variants~\cite{iia,chefer2021mm}. Perturbation-based methods derive the influence of input regions by introducing perturbation and analyzing their impact on model output. They often apply a mask to the input image and formulate an optimization problem to identify the most influential regions.
Another line of work aligns the model attention with human attention~\cite{xianyu:2021:scanpath,vqahat,sood2023multimodal} for better grounding. However, deep networks often make decisions with mechanisms different from humans, hence human attention is not very relevant for analyzing models. Different from them, we analyze the intrinsic behavior of LVLMs.
Traditional model interpretation methods are still limited to interpreting a single output, and can not directly apply to free-form generative models. Besides, current LVLMs often involve complex multi-modal structures with multi-resolution and multiple vision encoders, which makes the gradient-based and transformer interpretation methods~\cite{chefer2021transformer,hao2021self} hard to apply.

\noindent\textbf{LVLM Interpretation.}
Recently the interpretation of LVLMs has drawn increasing interest, as it provides valuable guidance for model development. A pipeline~\cite{ben2024lvlm} has been proposed to visualize the attention of LLaVA~\cite{llava}. Besides, some works visualize the model attention across different layers and either propose to improve the efficiency of LVLMs by pruning the redundant image tokens~\cite{chen2025image,zhang2024redundancy}, or alleviate the hallucination problem through modifying the decoding process~\cite{huang2024opera}. However, these methods mainly interpret the LVLMs from internal attention. The relevance of different regions on the input image to the output remains under-explored. A recent work~\cite{giulivi2024explaining} combines an open-world localization model with the LVLM to generate object localization for the output tokens using the vision embedding, while still limited to object-centric interpretation.

%% file: 3_method.tex
\vspace{-0.2cm}
\section{Method}
\label{sec:method}
\vspace{-0.2cm}
In this section, we first formulate the task and introduce some background knowledge about optimization-based heatmap visualization. Next, we propose a visually relevant token selection strategy and generalize the visualization method to open-ended responses of LVLMs with several improvements.
\subsection{Preliminaries}
\label{sec:preliminary}
\vspace{-0.1cm}
\noindent\textbf{Task Formulation.}
Given an input image $I$ and question $Q$ about the image, the LVLM generates an open-ended answer $a$ in natural language. 
The answers are generated autoregressively and may vary in length. In autoregressive text generation, words are tokenized and sequentially predicted conditioned on previous tokens. Suppose the answer $a$ consists of $l$ tokens, represented as a sequence $a=\left[a_1, a_2\cdots a_l\right]$. At each step $t$, the model samples the next token $a_t$ according to:
\vspace{-0.2cm}
\begin{align}
    a_t\sim P(a_t|a_1, a_2\cdots a_{t-1};I,Q)
\vspace{-0.1cm}
\end{align}

To investigate where the model focuses while generating the answer, we aim to obtain a heatmap $M$ that highlights the importance of each image pixel to the model output. In the next section we introduce the optimization-based heatmap visualization method iGOS++~\cite{khorram2021igos++}.

\noindent\textbf{Optimization-based Heatmap Visualization} derive the heatmap $M$ by solving an optimization problem. 
Suppose the model predicts a score $f$ for the output given $I, Q$, the optimization has two main objectives: \emph{deletion} and \emph{insertion}. \emph{Deletion} progressively removes pixels from $I$ in the order of their heatmap values, aiming to minimize the model's prediction score $f$. \emph{Insertion} starts with a baseline image $\Tilde{I}$ without visual information (\eg fully blurred image) and gradually restores pixels according to their heatmap values, optimizing the heatmap to maximize $f$. Hence the resulting heatmap highlights the most influential regions that contribute to the model’s final prediction.
The insertion and deletion operation can be denoted as:
\begin{align}
     \Phi(I, \Tilde{I}, M) = I \odot M + \tilde{I} \odot (1 - M)
\end{align}
where $\odot$ denotes the Hadamard product.
Defining deletion and insertion masks $M_x$ and $M_y$ for each objective, the final heatmap $M$ is obtained as their combination: $M = M_x \odot M_y$. The whole objective function is as follows:
\vspace{-0.1cm}
\begin{align}
    \min_{M = (M_x, M_y)}\;& f(\Phi(I, \tilde{I}, M_x)) - f(\Phi(I, \tilde{I}, 1 - M_y)) \notag \\
    +&f(\Phi(I, \tilde{I}, M)) - f(\Phi(I, \tilde{I}, 1 - M)) + g(M) \notag \\
    \text{where}\; &g(M) = \lambda_1 \|1 - M\|_1 + \lambda_2 BTV(M)
\vspace{-0.1cm}
\end{align}
$g(M)$ is a regularization term consisting of an $L1$ norm to promote sparsity and a Bilateral Total Variation (BTV) norm~\cite{khorram2021igos++} to enforce smoothness. The objective minimizes the deletion scores when applying $M_x$ and $M$, and maximizes the insertion scores with $M_y$ and $M$.

However, existing heatmap visualization methods cannot directly apply to the open-ended responses of LVLMs, since these models do not inherently produce a prediction score.
Furthermore, we find that this optimization process can be simplified by jointly optimizing a single heatmap for both deletion and insertion objectives, and the stability of the non-convex optimization can also be improved. We will discuss how to generalize this approach to open-ended responses in Section~\ref{sec:selection} and our improvements to the optimization method in Section~\ref{sec:adaptation}.

\subsection{Visually Relevant Token Selection}
\vspace{-0.1cm}
\label{sec:selection}
 \begin{figure}[t]
    \centering
    \includegraphics[width=0.45\textwidth]{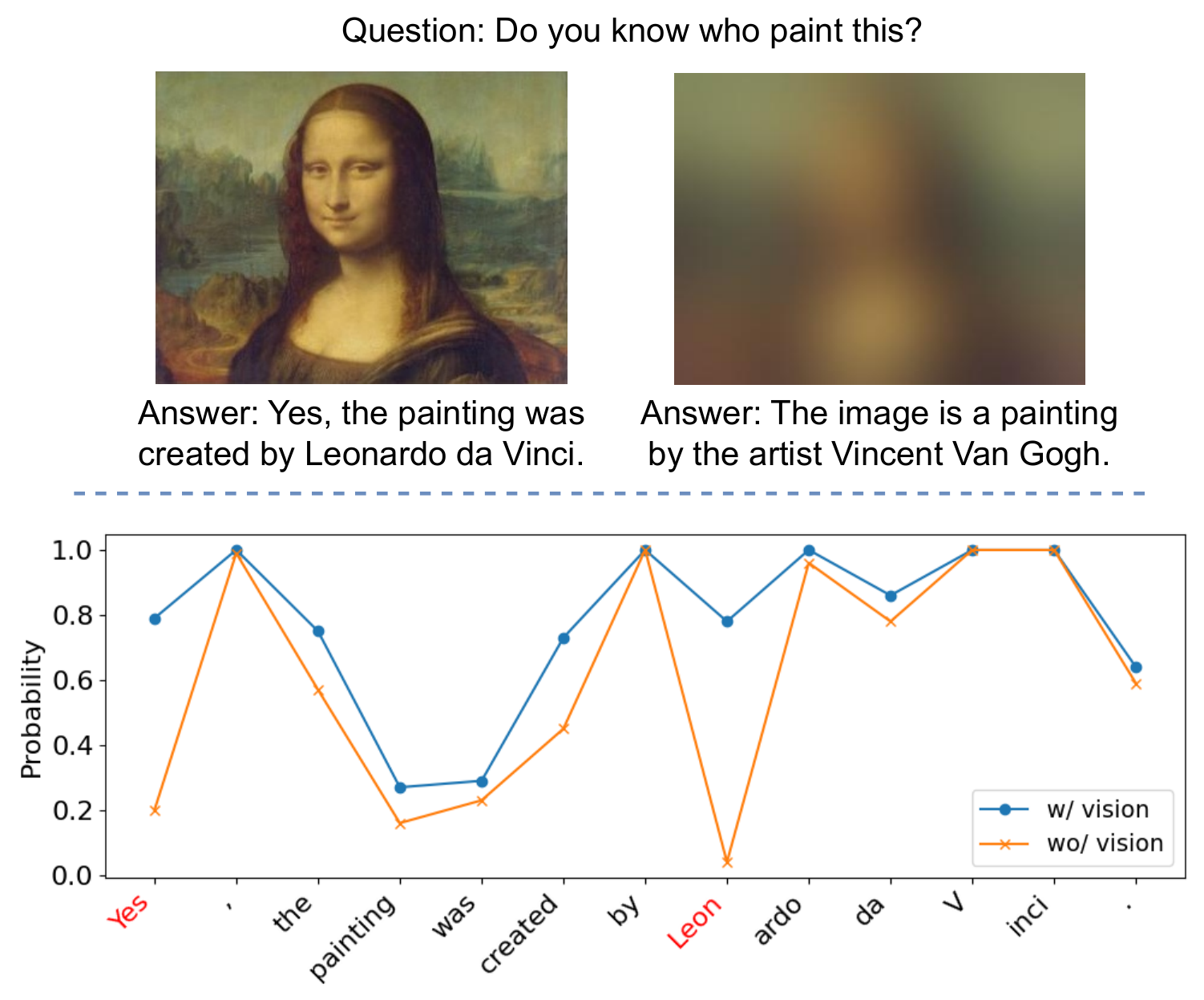} 
    \vspace{-0.4cm}
    \caption{Top figure: answers generated by LLaVA-1.5 given the original image and fully blurred baseline image. Bottom figure: conditional probability of the original answer given input image and baseline image. Most tokens in the response are not very dependent on the visual information.}
    \vspace{-0.4cm}
    \label{fig:token_prob}
\end{figure}
To extend heatmap visualization methods to free-form text outputs, a representative prediction score $f$ is required for optimization. A simple way to obtain a single score is by averaging all token probabilities. However, autoregressive text generation produces responses of variable length, where token-image correlations may vary significantly.
Specifically, token probabilities are strongly influenced by position within the sentence and the word. At the sentence level, certain words can be inferred based on syntax, making them less informative for assessing visual relevance. Within a word, subword token probabilities also vary significantly. As illustrated in Figure~\ref{fig:token_prob},
the conditional probabilities of certain tokens remain largely unchanged when the input is blurred, particularly for those that can be inferred from sentence structure and context (\eg, punctuation, painting, was, by) and the subsequent tokens within a word (\eg ardo, da, V, inci). 
In contrast, the first token in the sentence and some visually relevant tokens often exhibit a notable probability drop, as the first token in the response often decide the subsequent sentence structure, and the visually relevant tokens highly depend on the specific visual input. 
To this end, we propose to extract the most visually relevant tokens and derive the prediction score from them to achieve a representative interpretation of the whole output.

We formulate the relevance of the output tokens to the input image as the log-likelihood ratio~\cite{barnett2012transfer,woolf1957log} between the prediction with and without visual information. 
Due to the autoregressive nature of LVLMs, the probability of generating answer $a$ can be decomposed as the joint probability of its tokens. For simplicity, we omit $I$, $Q$ and denote the conditional probability of the next token as $P(a_t|a_1\cdots a_{t-1})$. The probability of the whole sentence is then:
\vspace{-0.1cm}
\begin{align}
P(a|I,Q)=P(a_1)P(a_2|a_1)\cdots P(a_l|a_1\cdots a_{l-1})
\end{align}

To measure the influence of visual information, we introduce a baseline image $\Tilde{I}$ that does not provide any visual information to answer the question. The probability of generating the original answer $a$ given $\Tilde{I}$ is denoted as: $P(a|\Tilde{I},Q)=\Tilde{P}(a_1)\cdots \Tilde{P}(a_l|a_1\cdots a_{l-1})$, where $\Tilde{P}(a_t|a_1\cdots a_{t-1})$ represents $P(a_t|a_1\cdots a_{t-1};\Tilde{I},Q)$. This term can be efficiently computed in a single forward pass by concatenating $Q$ and $a$ as the textual prompt to the model. The log-likelihood ratio (LLR) quantifies the difference in answer confidence with and without visual information:
\begin{align}
\label{eq:llr}
    &LLR=\log P(a|I,Q)-\log P(a|\Tilde{I},Q) \\
    &= \log \prod_{t}P(a_t|a_1 \cdots a_{t-1})-\log \prod_{t}\Tilde{P}(a_t|a_1 \cdots a_{t-1}) \notag \\
    &= \sum_t \log P(a_t|a_1 \cdots a_{t-1})- \sum_t \log \Tilde{P}(a_t|a_1 \cdots a_{t-1}) \notag
    \vspace{-0.2cm}
\end{align}
It can be observed that LLR of the whole output is the sum of $LLR_t$ of each token $a_t$. Therefore, to identify tokens most influenced by visual information, we apply a threshold to filter those with the highest log-likelihood ratio. 
The set of crucial tokens $\mathcal{K}$ is selected as:
\begin{align}
   \mathcal{K} =\{ a_k &| \, LLR_k > \alpha; \;k\neq 1\}
\end{align}

Finally, we define the prediction score $f$ as the cumulative log-likelihood of the crucial tokens. It ensures that only visually relevant parts of the response contribute to the interpretation, filtering out the influence of linguistic structures and leading to a more faithful interpretation of the model’s reliance on visual information.
\vspace{-0.1cm}
\begin{align}
    f=\sum_{a_k\in \mathcal{K}} \log P(a_k | a_1 \dots a_{k-1})
\vspace{-0.1cm}
\label{eq:score}
\end{align}

\subsection{Adaptation to LVLMs}
\label{sec:adaptation}
\noindent\textbf{Non-convex Optimization.}
In practice, the optimization process is non-convex and difficult to converge. Besides, the reasoning process of LVLMs is complicated, with responses often related to multiple regions in the image, leading to scattered attention maps that are hard to optimize. To address this, we simplify the objective function by directly optimizing a single mask $M$ for both the deletion and insertion objectives. To further improve stability, 
we incorporate graduated non-convexity~\cite{hazan2016graduated} (GNC) to reduce the risk of getting into local optima and the oscillations during optimization.
Instead of directly solving a highly non-convex problem, GNC begins with a convex approximation to provide a more stable starting point, and gradually introduces non-convexity into the optimization process.
Specifically, we add an exponentially decayed $L2$ norm to the objective function, yielding the final formulation:
\begin{align}
\label{eq:objective}
    &\min_{M}\;f(\Phi(I, \tilde{I}, M)) - f(\Phi(I, \tilde{I}, 1 - M)) + g(M) \\
&g(M) = \lambda_1 \|1 - M\|_1 + \lambda_2 e^{-\gamma t} \|1 - M\|_2 + \lambda_3 BTV(M) \notag
\end{align}
and we set $\lambda_1=1, \lambda_2=0.1, \lambda_3=10$ as the default value. 

\noindent\textbf{Multi-encoder and Multi-resolution.}
Current LVLMs often leverage multiple vision encoders~\cite{cambrian} or multi-resolution~\cite{llavaov,li2024mini} input images to enhance visual understanding, which introduces challenges when applying the mask to the input image. For the multi-encoder architectures, the input image is processed by multiple vision encoders before integrating their features. To accommodate this, we apply a single unified mask to the input image before passing it through all encoders, ensuring consistency across different feature extractors. For the multi-resolution methods, they often crop the input image into variable-sized patches according to the original resolution, where the operation is nondifferentiable and obstructs the optimization. To overcome this, we implement an equivalent differentiable cropping operation (replace pillow and numpy operations with tensor operations), ensuring that the mask undergoes the same transformation as the image patches. This allows us to apply the multi-resolution mask to the corresponding image patches while maintaining differentiability.

%% file: 4_experiment.tex
\vspace{-0.1cm}
\section{Experiment}
\label{sec:experiment}
\vspace{-0.1cm}
We leverage our improved heatmap visualization method to interpret the open-ended output of LVLMs. To gain insights into the visual behavior of state-of-the-art open-source LVLMs, we conduct quantitative and qualitative experiments, focusing on the following key research questions. \textbf{Q1}: Do LVLMs rely on the input image when answering visual questions? \textbf{Q2}: Where do different LVLMs attend when generating variable-length responses? \textbf{Q3}: What is the relationship between answer correctness and focus region? \textbf{Q4}: How do the vision encoder and LLM components influence visual behavior?

\begin{table*}[!t]
\centering
\begin{tabular}{clcccccccc}
\hline
\multicolumn{1}{l}{} &  & \multicolumn{2}{c}{MMVP} & \multicolumn{2}{c}{MMStar} & \multicolumn{2}{c}{CV-Bench} & \multicolumn{2}{c}{LLaVA-Bench} \\ \hline
\multicolumn{1}{l}{LVLM} & Method & Del $\downarrow$ & Ins $\uparrow$ & Del $\downarrow$ & Ins $\uparrow$ & Del $\downarrow$ & Ins $\uparrow$ & Del $\downarrow$ & Ins $\uparrow$ \\ \hline
\multirow{4}{*}{\begin{tabular}[c]{@{}c@{}}LLaVA-1.5-7b\end{tabular}} & Grad-CAM & 0.679 & 0.441 & 0.689 & 0.372 & 0.651 & 0.587 & 0.685 & 0.379 \\
 & T-MM & 0.778 & 0.418 & 0.869 & 0.283 & 0.869 & 0.461 & 0.808 & 0.378 \\
 & IIA & 0.401 & 0.805 & 0.333 & 0.870 & 0.457 & 0.884 & 0.371 & 0.824 \\
 & Ours & \textbf{0.366} & \textbf{0.811} & \textbf{0.292} & \textbf{0.953} &\textbf{0.402} & \textbf{0.965} & \textbf{0.358} & \textbf{0.864} \\ \hline
\multirow{4}{*}{\begin{tabular}[c]{@{}c@{}}LLaVA-OV-7b\end{tabular}} & Grad-CAM & 0.341 & 0.501 & 0.406 & 0.550 & 0.443 & 0.562 & 0.334 & 0.549 \\
 & T-MM & 0.528 & 0.589 & 0.575 & 0.620 & 0.594 & 0.630 & 0.589 & 0.655 \\
 & IIA & 0.576 & 0.551 & 0.621 & 0.578 & 0.660 & 0.553 & 0.541 & 0.612 \\
 & Ours & \textbf{0.305} & \textbf{0.778} & \textbf{0.317} & \textbf{0.870} & \textbf{0.295} & \textbf{0.924} & \textbf{0.301} & \textbf{0.803} \\ \hline
 \multirow{2}{*}{\begin{tabular}[c]{@{}c@{}}Cambrian-8b\end{tabular}} & Grad-CAM & 0.416 & 0.513 & 0.391 & 0.656 & 0.471 & 0.599 & 0.452 & 0.582 \\
 & Ours & \textbf{0.375} & \textbf{0.657} & \textbf{0.334} & \textbf{0.860} & \textbf{0.340} & \textbf{0.849} & \textbf{0.372} & \textbf{0.786} \\ \hline
\end{tabular}
\vspace{-0.2cm}
\caption{Quantitative comparison of interpretation methods in terms of Deletion score (lower is better), Insertion score (higher is better) on the filtered dataset using different LVLMs. The bold numbers denote the best results for each model and dataset.}
\vspace{-0.2cm}
\label{tab:comparison}
\end{table*}

\subsection{Models and Datasets}
We evaluate LVLMs that employ representative strategies to improve visual instruction following capabilities: LLaVA-1.5~\cite{liu2024improved} leverages a fully connected cross-modal adapter and incorporates academic-related data~\cite{vqav2} to enhance visual instruction tuning. LLaVA-OneVision~\cite{llavaov} employs multi-resolution input images, hence captures finer image details. Cambrian~\cite{cambrian} proposes a Spatial Vision Aggregator to integrate visual features from multiple encoders.
Given these architectural differences, we analyze how they affect the visual behaviors.

\vspace{-0.1cm}
We select recent datasets that target at evaluating the visual instruction following capabilities of LVLMs: MMStar~\cite{mmstar} contains 1,500 human-reviewed vision-dependent questions that most LVLMs fail to answer correctly without visual input. CV-Bench~\cite{cambrian} constructs a ``vision-centric'' benchmark with 2,638 manually verified image-related questions. MMVP~\cite{mmvp} selects a subset of 300 questions with the images that CLIP~\cite{clip} fails to distinguish.

\begin{table}[tb]
\vspace{-0.2cm}
\begin{tabular}{lccc}
\hline
          & CV-Bench    & MMStar      & MMVP        \\ \hline
LLaVA-1.5 & 8.9 / 18.5 & 7.5 / 10.4 & 25.3 / 16.3 \\
LLaVA-OV  & 13.6 / 20.2 & 4.7 / 5.7  & 16.3 / 15.0 \\
Cambrian  & 1.1 / 2.3   & 4.2 /  5.0 & 3.3 / 7.3  \\ \hline
\end{tabular}
\vspace{-0.2cm}
\caption{Percentage ($\%$) of samples that the answer probability decreases by less than 30$\%$ / 30$\%$-70$\%$ without visual information.}
\vspace{-0.5cm}
\label{tab:ratio}
\end{table}

\vspace{-0.2cm}
\subsection{Statistical Analysis}
\label{statistics}
\vspace{-0.1cm}
To investigate \textbf{Q1} (whether LVLMs rely on visual input), we conduct a statistical analysis comparing the models’ responses with and without visual information.
Using the visual relevance metric in Eq.~\ref{eq:llr}, 
 we compute the probability of an answer given the original image versus a fully blurred image. Table~\ref{tab:ratio} reports the percentage of samples where the answer probability decreases by less than 30$\%$ or between 30$\%$-70$\%$ without visual information. The results indicate that most responses are affected by the image to varying degrees.
Notably, LLaVA-1.5 exhibits lower reliance on visual input on MMStar and MMVP; with 25.3$\%$ MMVP samples showing a probability drop of less than 30$\%$ when the image is blurred.
 The responses of Cambrian are more influenced by the visual contents; with answer probabilities decreasing by more than $70\%$ for $\sim90\%$ of samples across datasets. 
 Most compared models have lower density of small probability drops on MMStar, suggesting they rely more heavily on image when answering MMStar questions.

\vspace{-0.2cm}
\subsection{Comparison of Visualization Methods}
\vspace{-0.1cm}
Our proposed token selection method can be applied to various interpretation methods and extend them to open-ended responses of LVLMs. However, some previous interpretation techniques are not well-suited for LVLMs.
To demonstrate the advantage of our proposed method over other heatmap visualization methods, we compare the gradient-based method Grad-CAM~\cite{gradcam}, transformers interpretation methods T-MM~\cite{chefer2021mm} and IIA~\cite{iia}, and our improved optimization-based method based on IGOS++~\cite{khorram2021igos++}.

\noindent\textbf{Evaluation Metric.}
We follow the commonly used \emph{deletion} and \emph{insertion}~\cite{petsiuk2018rise} scores to assess the heatmaps. Deletion removes pixels from the original image in descending order of their heatmap values, and calculates the output scores given the intermediate image to derive a deletion curve. The deletion score is the area under the curve (AUC). Similarly, the insertion score measures how quickly the output score increases when adding pixels to a baseline image.
Lower deletion score and higher insertion score indicate heatmap that better reflects areas the model attends to. 
For fair comparison across models with varying prediction score distributions, we normalize the scores according to those of the original image and baseline image, following~\cite{jiang2024comparing}.

\noindent\textbf{Data Selection.}
For subsequent studies, we filter out the samples where models' answers are largely independent of the image, as it is infeasible to study where the model attends if it does not need the image. We keep the samples where all models have clear probability differences with and without visual information, remaining $35\%$ samples from MMVP, $47\%$ of MMStar samples, and $34\%$ of CV-Bench. The original datasets are multiple-choice questions, so we remove the choices and instructions to relax them into open-ended questions. Additionally, we evaluate on LLaVA-Bench~\cite{llava} designed for open-ended VQA.

\begin{figure}[t]
    \captionsetup{font=small}
    \captionsetup{justification=centering}
    \centering
    \begin{subfigure}{0.47\textwidth}
        \centering
        \caption*{\small Question: \textit{What are the words in the image?} \\ LLaVA-1.5: \textit{The \textcolor{red}{words} in the image are "\textcolor{red}{Happy Easter}."}}
        
        \begin{minipage}{0.19\textwidth}
            \includegraphics[width=\textwidth]{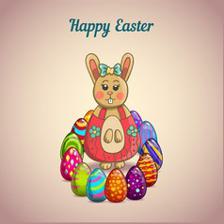}
            \caption*{\scriptsize Input image}
        \end{minipage}
        \begin{minipage}{0.19\textwidth}
            \includegraphics[width=\textwidth]{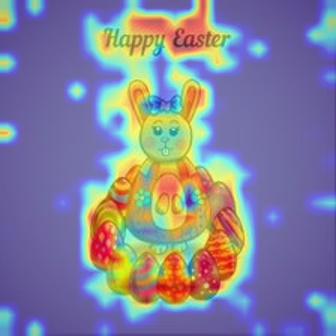}
            \caption*{\scriptsize Grad-CAM}
        \end{minipage}
        \begin{minipage}{0.19\textwidth}
            \includegraphics[width=\textwidth]{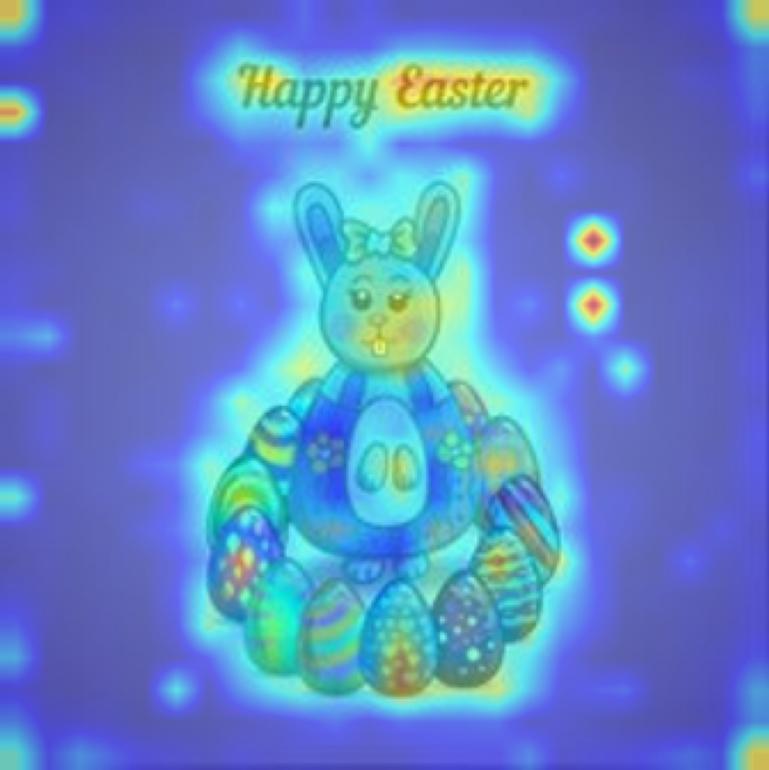}
            \caption*{\scriptsize T-MM}
        \end{minipage}
        \begin{minipage}{0.19\textwidth}
            \includegraphics[width=\textwidth]{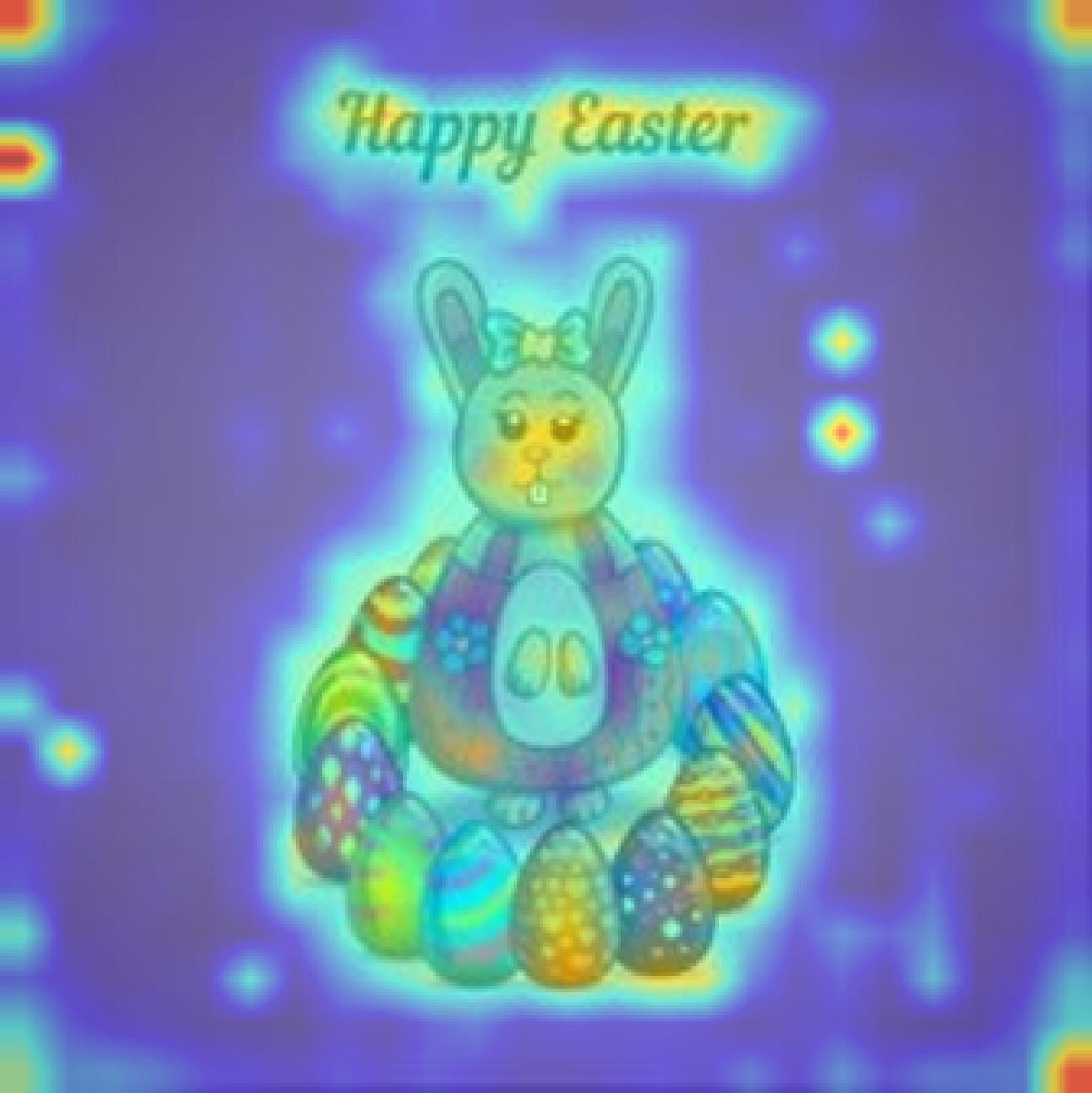}
            \caption*{\scriptsize IIA}
        \end{minipage}
        \begin{minipage}{0.19\textwidth}
            \includegraphics[width=\textwidth]{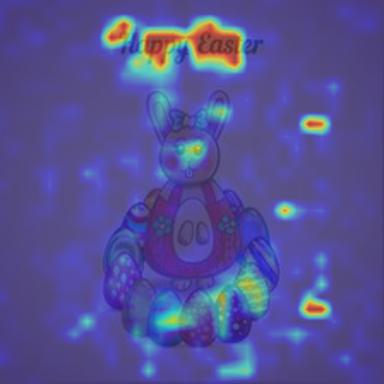}
            \caption*{\scriptsize Ours}
        \end{minipage}
    \end{subfigure}

    \begin{subfigure}{0.47\textwidth}
        \centering
        \caption*{\small Question: \textit{Are the ears of the dog erect or drooping?} \\ LLaVA-OV: \textit{The ears of the dog are \textcolor{red}{erect}.}}

        \begin{minipage}{0.19\textwidth}
            \includegraphics[width=\textwidth]{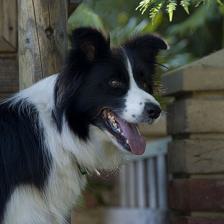}
            \caption*{Input image}
        \end{minipage}
        \begin{minipage}{0.19\textwidth}
            \includegraphics[width=\textwidth]{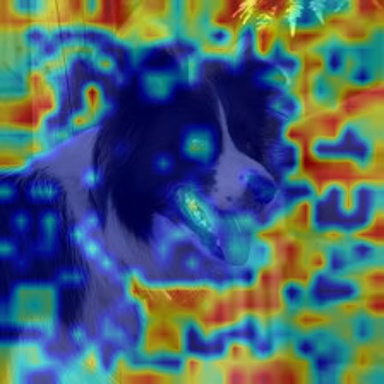}
            \caption*{Grad-CAM}
        \end{minipage}
        \begin{minipage}{0.19\textwidth}
            \includegraphics[width=\textwidth]{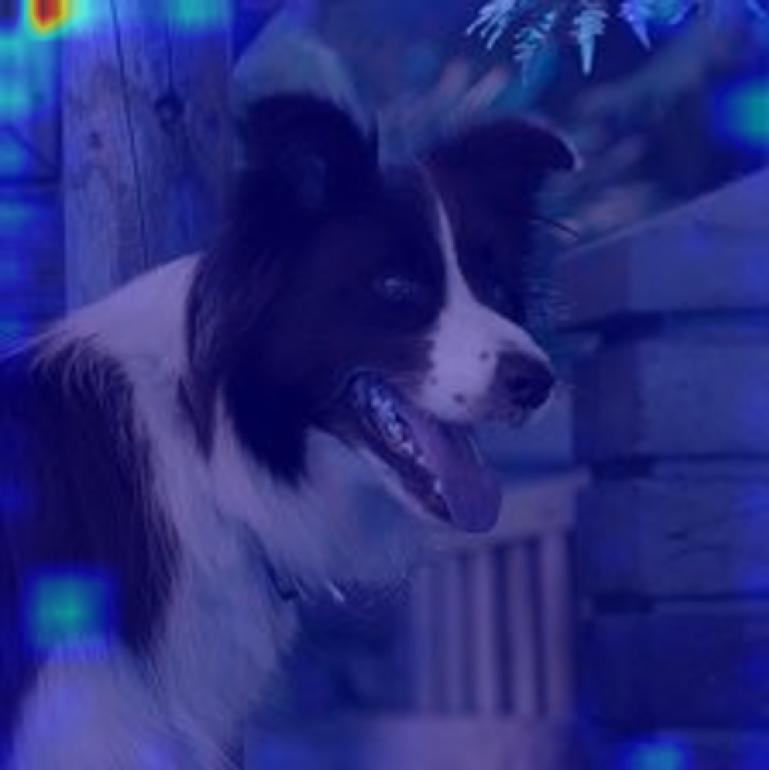}
            \caption*{T-MM}
        \end{minipage}
        \begin{minipage}{0.19\textwidth}
            \includegraphics[width=\textwidth]{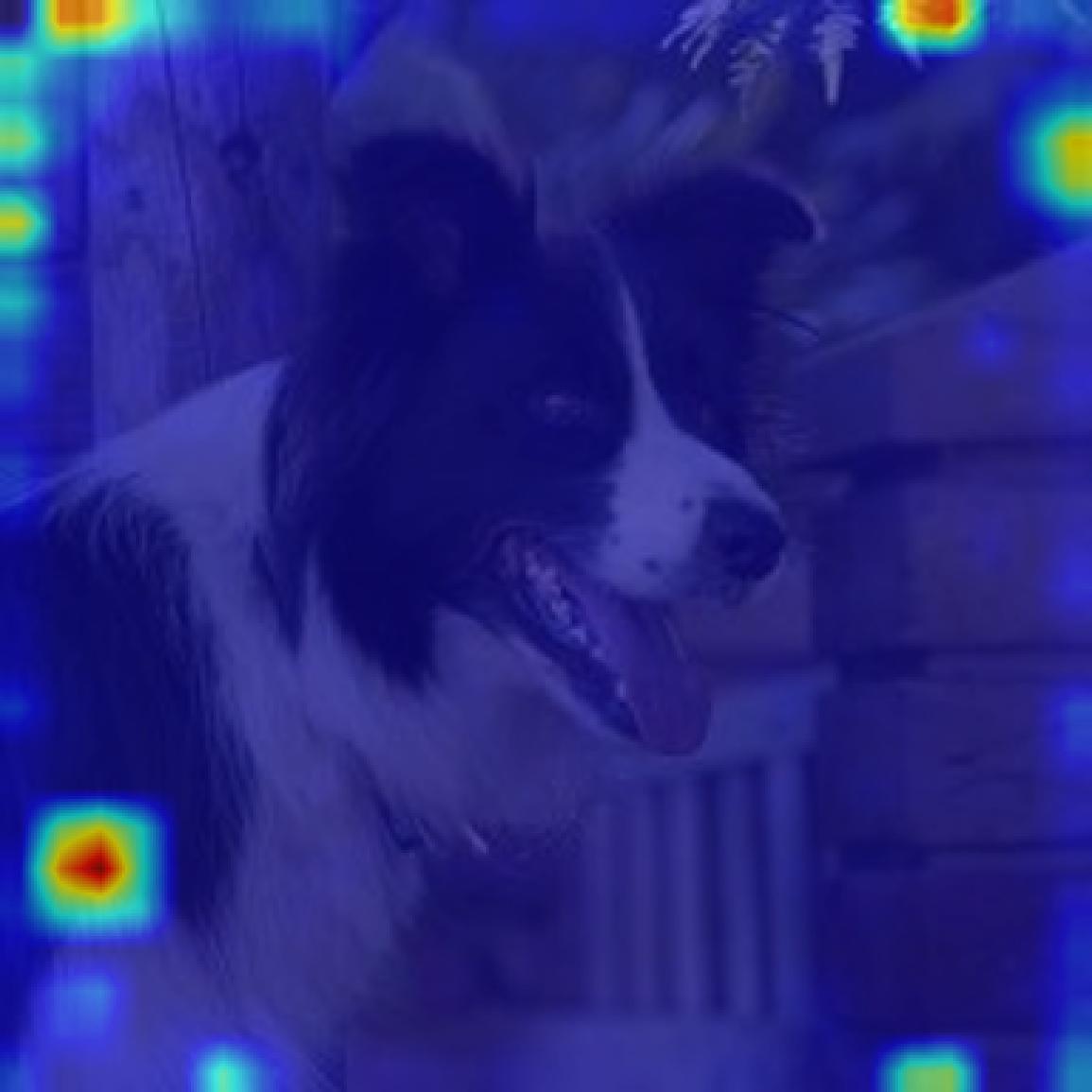}
            \caption*{\scriptsize IIA}
        \end{minipage}
        \begin{minipage}{0.19\textwidth}
            \includegraphics[width=\textwidth]{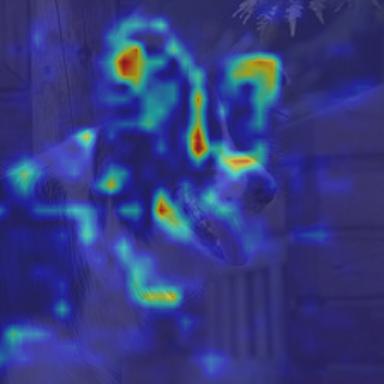}
            \caption*{\scriptsize Ours}
        \end{minipage}

    \end{subfigure}
    \vspace{-0.2cm}
    \captionsetup{justification=justified}
    \caption{Qualitative comparison of different explanation methods. The tokens in red denote the selected crucial tokens. Our method consistently generates meaningful heatmaps.}
    \vspace{-0.4cm}
    \label{fig:comparison}
\end{figure}

\noindent\textbf{Experiment Results.}
With the selected data, we generate heatmaps using different interpretation methods and compare their deletion and insertion scores in Table~\ref{tab:comparison}. As Cambrian uses multiple vision encoders including non-transformer-based models, T-MM and IIA are not applicable. Our method consistently outperforms all baselines, with the lowest deletion scores and highest insertion scores across datasets and models. We further compare qualitative heatmap visualizations in Figure~\ref{fig:comparison}, showing that our method generates more meaningful heatmaps and adapts better to diverse model structures. These results ensure the reliability of our method for the subsequent analysis. 

\subsection{Ablation Study}
\vspace{-0.1cm}
We conduct ablation studies on the key components of our method including the visual relevant token selection, baseline image choice and single-mask optimization. The results are presented in Table~\ref{tab:ablation}, where \emph{Proposed} refers to our default configuration with the proposed token selection strategy, using blurred image as the baseline image, and optimizing single mask with GNC. Additional parameter studies are provided in Appendix \ref{sec:app_ablation}.

\noindent\textbf{Baseline Image.}
The baseline image $\Tilde{I}$ should contain minimal visual information~\cite{fong2017interpretable}, while maintaining a distribution consistent with natural images to avoid introducing adversarial artifacts. We compare different choices of the baseline image in Table~\ref{tab:ablation}, including blurred image, all-zero input, and random noise.
Results indicate that the blurred baseline image achieves the best balance, minimally disturbing the input image distribution.

\noindent\textbf{Visually Relevant Tokens.} 
An important contribution of our method is handling long-sentence responses by identifying crucial tokens based on the log-likelihood ratio. To demonstrate its effectiveness, we compare the token selection strategy with (1) Computing the joint probability of the whole sentence. (2) Detecting keywords using off-the-shelf tagging method~\cite{campos2018text,campos2020yake}. As shown in Table~\ref{tab:ablation}, our proposed method achieves overall best results of deletion and insertion scores. In comparison, computing the joint probability may introduce irrelevant tokens that do not reflect visual information. Moreover, the keywords detection method operates at the word level instead of the token level, which does not align well with the tokenized output of LLM. 

\noindent\textbf{Single Mask Optimization.}
Next we evaluate the efficacy of replacing separate masks for deletion and insertion with a single mask for optimization. As shown in Table~\ref{tab:ablation}, using separate masks provides slightly better scores. However, it significantly increases computation time, with an average runtime per sample of 19.1s, compared to only 7.4s for the single-mask approach. The visualized results show that optimizing a single mask can also yield reasonable results. Therefore, our simplified optimization achieves better balance between performance and efficiency.
\begin{table}[tbp]
\resizebox{0.47\textwidth}{!}{
\begin{tabular}{lcccccc}
\hline
 & \multicolumn{2}{c}{MMVP} & \multicolumn{2}{c}{MMStar} & \multicolumn{2}{c}{CV-Bench} \\ \hline
 & Del $\downarrow$ & Ins $\uparrow$ & Del $\downarrow$ & Ins $\uparrow$ & Del $\downarrow$ & Ins $\uparrow$ \\ \hline
Proposed & \textbf{0.366} & \textbf{0.811} & 0.292 & \textbf{0.953} & \textbf{0.402} & 0.965 \\ \hline
\multicolumn{7}{c}{Comparison of output token selection strategy} \\ \hline
Joint prob. & 0.403 & 0.784 & \textbf{0.286} & 0.906 & 0.409 & \textbf{0.967} \\
Keywords & \textbf{0.367} & 0.809 & 0.302 & 0.931 & \textbf{0.402} & 0.899 \\ \hline
\multicolumn{7}{c}{Comparison of different baseline image selection} \\ \hline
Blank & 0.407 & 0.644 & 0.317 & 0.805 & 0.403 & 0.837 \\
Noise & 0.381 & 0.696 & 0.295 & 0.838 & 0.392 & 0.884 \\ \hline
\multicolumn{7}{c}{Comparison with separate masks for insertion/deletion.} \\ \hline
Separate & 0.373 & \textbf{0.819} & \textbf{0.279} & \textbf{0.965} & \textbf{0.383} & \textbf{0.975} \\ \hline
\end{tabular}
}
\caption{Ablation study of visually relevant token selection, the baseline image and simplifying separate masks to single mask. The bold numbers denote the best and second-best results among the compared variants.}
\vspace{-0.4cm}
\label{tab:ablation}
\end{table}

\begin{figure*}[tb]
    \captionsetup{justification=centering}
    \centering
    \captionsetup{font=small}
    \begin{subfigure}[t]{\columnwidth}
        \centering
        \caption*{(a) Question: \textit{What color is the chicken's body?}}
        \begin{minipage}[t]{0.24\textwidth}
            \includegraphics[width=\textwidth]{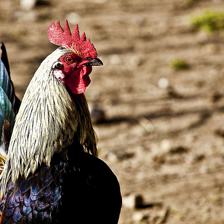}
            \caption*{\scriptsize Input image}
        \end{minipage}
        \begin{minipage}[t]{0.24\textwidth}
            \includegraphics[width=\textwidth]{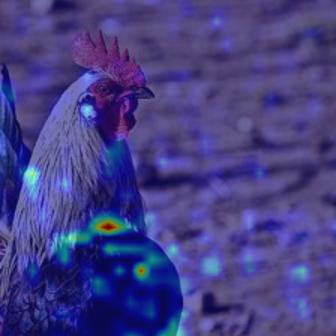}
            \caption*{\scriptsize LLaVA-1.5\\ \textit{The chicken's body is \textcolor{red}{black} and white.}}
        \end{minipage}
        \begin{minipage}[t]{0.24\textwidth}
            \includegraphics[width=\textwidth]{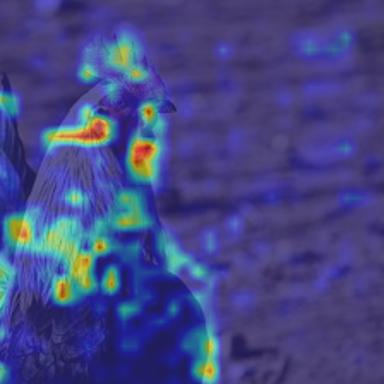}
            \caption*{\scriptsize LLaVA-OV \\ \textit{\textcolor{red}{black}.}}
        \end{minipage}
        \begin{minipage}[t]{0.24\textwidth}
            \includegraphics[width=\textwidth]{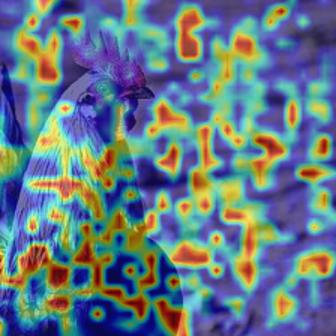}
            \caption*{\scriptsize Cambrian \\ \textit{The chicken's body is primarily \textcolor{red}{black}, with some \textcolor{red}{ir}idescent blue and green feathers.}}
        \end{minipage}

        \caption*{(c) Question: \textit{Do the individuals in the picture face the front or the back?}}
        \begin{minipage}[t]{0.24\textwidth}
            \includegraphics[width=\textwidth]{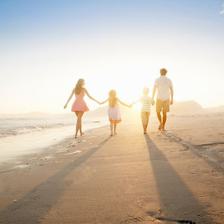}
            \caption*{\scriptsize Input image}
        \end{minipage}
        \begin{minipage}[t]{0.24\textwidth}
            \includegraphics[width=\textwidth]{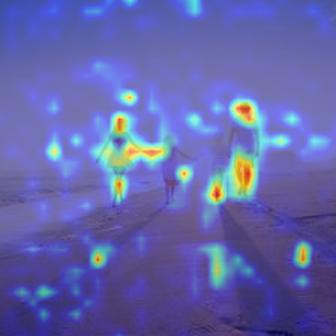}
            \caption*{\scriptsize LLaVA-1.5\\ \textit{The individuals in the picture face the front, as they are \textcolor{red}{walking together} on the beach.}}
        \end{minipage}
        \begin{minipage}[t]{0.24\textwidth}
            \includegraphics[width=\textwidth]{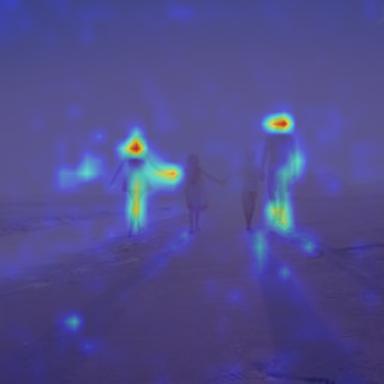}
            \caption*{\scriptsize LLaVA-OV \\ \textit{The individuals in the picture are facing \textcolor{red}{away} from the camera, \textcolor{red}{looking} towards the \textcolor{red}{horizon}.}}
        \end{minipage}
        \begin{minipage}[t]{0.24\textwidth}
            \includegraphics[width=\textwidth]{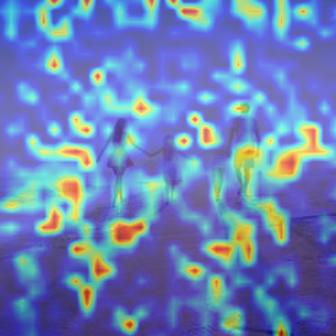}
            \caption*{\scriptsize Cambrian \\ \textit{The individuals in the picture are facing the front, \textcolor{red}{moving} towards the \textcolor{red}{horizon}.}}
        \end{minipage}

    \end{subfigure}
    \begin{subfigure}[t]{\columnwidth}
    \centering

    \caption*{(b) Question: \textit{Can you see people in this image?}}
        \begin{minipage}[t]{0.24\textwidth}
            \includegraphics[width=\textwidth]{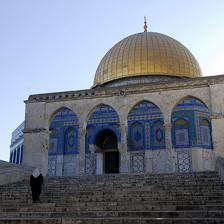}
            \caption*{\scriptsize Input image}
        \end{minipage}
        \begin{minipage}[t]{0.24\textwidth}
            \includegraphics[width=\textwidth]{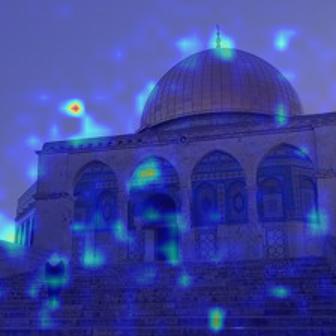}
            \caption*{\scriptsize LLaVA-1.5\\ \textit{Yes, there is a person visible in the image, standing on the \textcolor{red}{steps} in front of the building.}}
        \end{minipage}
        \begin{minipage}[t]{0.24\textwidth}
            \includegraphics[width=\textwidth]{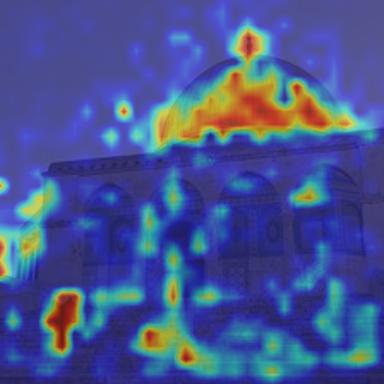}
            \caption*{\scriptsize LLaVA-OV \\ \textit{Yes, there is a person in the image. They appear to be \textcolor{red}{walking up} the \textcolor{red}{steps towards} the \textcolor{red}{building} with the \textcolor{red}{golden} dome.}}
        \end{minipage}
        \begin{minipage}[t]{0.24\textwidth}
            \includegraphics[width=\textwidth]{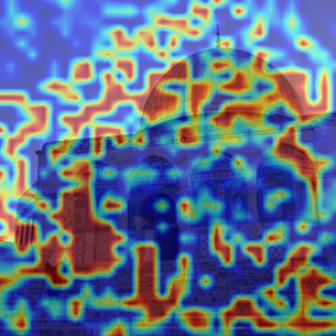}
            \caption*{\scriptsize Cambrian \\ \textit{Yes, there is a person visible in the image, \textcolor{red}{walking down} the \textcolor{red}{steps} of the \textcolor{red}{Al A}qsa Mosque.}}
        \end{minipage}

    \caption*{(d) Question: \textit{What color are the flowers in the background?}}
        \begin{minipage}[t]{0.24\textwidth}
            \includegraphics[width=\textwidth]{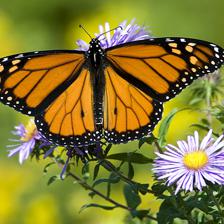}
            \caption*{\scriptsize Input image}
        \end{minipage}
        \begin{minipage}[t]{0.24\textwidth}
            \includegraphics[width=\textwidth]{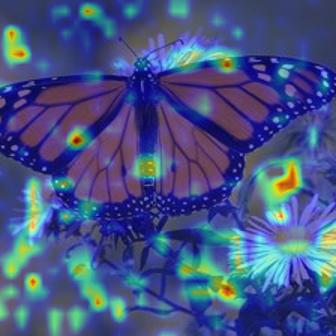}
            \caption*{\scriptsize LLaVA-1.5\\ \textit{The flowers in the background are \textcolor{red}{pur}ple.}}
        \end{minipage}
        \begin{minipage}[t]{0.24\textwidth}
            \includegraphics[width=\textwidth]{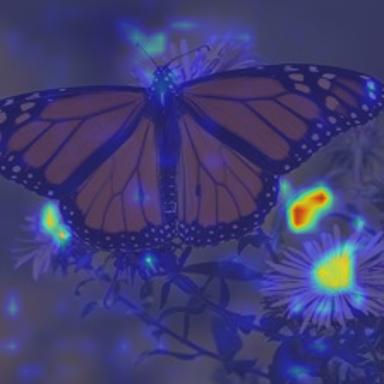}
            \caption*{\scriptsize LLaVA-OV \\ \textit{\textcolor{red}{yellow}.}}
        \end{minipage}
        \begin{minipage}[t]{0.24\textwidth}
            \includegraphics[width=\textwidth]{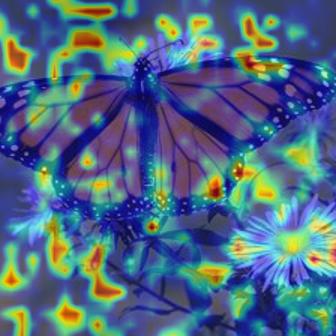}
            \caption*{\scriptsize Cambrian \\ \textit{The flowers in the background are \textcolor{red}{purple}.}}
        \end{minipage}
        
    \end{subfigure}

    \captionsetup{justification=justified}
    \vspace{-0.3cm}
    \caption{Comparison of the generated response and focus region of different LVLMs. Tokens in red are the selected visual relevant tokens.}
    \vspace{-0.2cm}
    \label{fig:visualization1}
\end{figure*}

 \begin{figure*}[!t]
    \centering
    \includegraphics[width=0.8\textwidth]{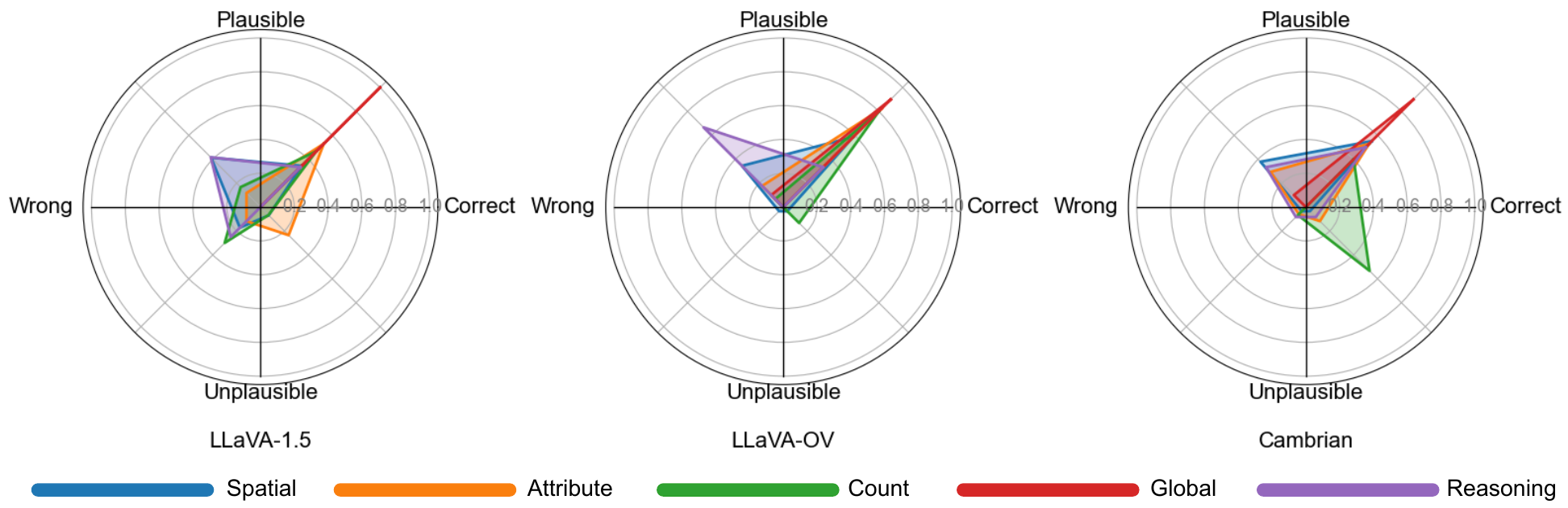}
    \vspace{-0.2cm}
    \caption{Answer correctness and focus region plausibility across four quadrants. Each color stands for a different question category, including spatial, attribute, counting, global context, and reasoning questions.}
    \label{fig:radar}
    \vspace{-0.4cm}
\end{figure*}

\subsection{Focus region analysis}
\vspace{-0.1cm}
\noindent\textbf{Heatmap Visualization.}
We present qualitative results in Figure~\ref{fig:visualization1} to address \textbf{Q2} about the focus region of different LVLMs when generating outputs. We highlight the visually relevant tokens in red. 
The results lead to several observations: (1) Cambrian reveals more compositional~\cite{jiang2024comparing} image understanding, which means it tends to jointly consider the entire image and involve more comprehensive information in its responses (\eg in example (a) Cambrian provides more information about the chicken's color). In contrast, LLaVA-OV shows more disjunctive behaviors and focuses on specific regions, performing better on questions requiring detailed understanding (\eg in example (c), only LLaVA-OV correctly determines whether the individuals are facing the front or back. 
This aligns well with their respective architectures: Cambrian aggregates multiple vision encoders to extract broader visual information, while LLaVA-OV adopts a multi-resolution strategy to extract detailed features. (2) The models' responses may involve contents not asked for in the questions but are related to some image regions. For instance, in Figure~\ref{fig:visualization1}(b) LLaVA-OV describes the golden dome as it attends to that region, while Cambrian mentions the landmark's name, likely due to its global scene attention. (3) When LVLMs give incorrect answers, their focus regions often reveal the underlying cause. In example (d), LLaVA-OV locates the pistil instead of the petal hence answers yellow for the color of the flower. More visualization results are included in Appendix~\ref{app_visualize}.

\noindent\textbf{Answer vs. Focus Region.}
To address \textbf{Q3}, we conduct a subjective analysis on the relationship between answer correctness and focus region plausibility (\emph{i.e.,} whether the focus region aligns with human intuition). We randomly select 100 samples for and categorize them into spatial, attribute, counting, global context, and reasoning questions. The results are summarized in Figure~\ref{fig:radar}, where we classify model behaviors into four quadrants and calculate the percentage of samples in each quadrant.
The key observations include: (1) All models tend to provide correct answers with plausible focus regions when answering global context questions. Conversely, models often fail to answer reasoning and spatial questions even when attending to the right regions. (2) LLaVA-1.5 has lower chance to locate the most relevant regions. LLaVA-OV has better focus plausibility on most question types, though it does not necessarily lead to better accuracy. Cambrian performs well on counting questions, yet often attends to regions that do not align with human intuition.
These findings suggest that focus region plausibility does not always correlate with answer correctness.
In Appendix~\ref{sec:app_human}, we provide additional experiments to compare the model focus and human attention.

\begin{table*}[!tbp]
\centering
\begin{tabular}{lcccc}
\hline
 Model    & LLM             & Vision Encoder        & Del$\downarrow$  & Ins$\uparrow$  \\ \hline
\multicolumn{5}{c}{Comparison of models with different LLMs}            \\ \hline
LLaVA-OV-0.5b~\cite{llavaov}         & qwen2-0.5b~\cite{yang2024qwen2}      & SigLIP~\cite{siglip}          & 0.313   & 0.792   \\
LLaVA-OV-7b           & qwen2-7b        & SigLIP          & 0.305   & 0.778   \\
LLaVA-OV-72b          & qwen2-72b       & SigLIP          & 0.304   & 0.754   \\ \hline
Cambrian-3b~\cite{cambrian}           & Phi-3-3.8B~\cite{abdin2024phi}      & Multi-encoder   & 0.424   & 0.664   \\
Cambrian-8b           & LLaMA3-8B~\cite{dubey2024llama}       & Multi-encoder   & 0.375   & 0.657   \\
Cambrian-13b          & Vicuna1.5-13B   & Multi-encoder   & 0.415   & 0.692   \\ \hline
\multicolumn{5}{c}{Comparison of models with different vision architectures} \\ \hline
LLaVA-1.5-7b & Vicuna1.5-7B~\cite{vicuna} & CLIP~\cite{clip}   & 0.366 & 0.811 \\
Mini-Gemini-7b~\cite{li2024mini}   & Vicuna1.5-7B    & CLIP-L        & 0.474   & 0.669   \\
Mini-Gemini-7b-HD     & Vicuna1.5-7B    & ConvNext-L       & 0.478   & 0.661   \\ \hline
Cambrian-13b & Vicuna1.5-13B & Multi-encoder & 0.415   & 0.692 \\
Mini-Gemini-13b       & Vicuna1.5-13B   & CLIP-L        &   0.473      & 0.674   \\
Mini-Gemini-13b-HD    & Vicuna1.5-13B   & ConvNext-L       &  0.471   &   0.671  \\ \hline
\end{tabular}
\caption{Comparison of models with different LLM scales and vision architectures. For Cambrian models, they use a combination of 4 vision encoders: CLIP ViT-L~\cite{clip}, SigLIP ViT~\cite{siglip}, OpenCLIP ConvNeXt-XXL~\cite{openclip} and DINOv2 ViT-L~\cite{dinov2}. For Mini-Gemini models, \emph{HD} denotes high resolution with an additional vision encoder of ConvNext-L.}
\vspace{-0.2cm}
\label{tab:scale}
\end{table*}

\begin{figure*}[!htb]
    \captionsetup{font=small}
    \captionsetup{justification=centering}
    \centering
    \begin{subfigure}{0.98\textwidth}
        \centering
        \caption*{Question: \textit{Is the duck floating?}}
        
        \begin{minipage}[t]{0.13\textwidth}
            \includegraphics[width=\textwidth]{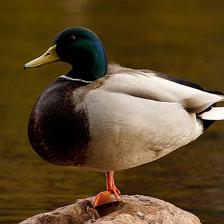}
            \caption*{\scriptsize Input image}
        \end{minipage}
        \begin{minipage}[t]{0.13\textwidth}
            \includegraphics[width=\textwidth]{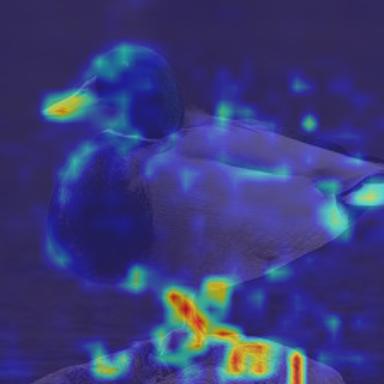}
            \caption*{\scriptsize LLaVA-OV-0.5b\\ \textit{No, the duck is not floating; it appears to be \textcolor{red}{standing} on a \textcolor{red}{rock}.}}
        \end{minipage}
        \begin{minipage}[t]{0.13\textwidth}
            \includegraphics[width=\textwidth]{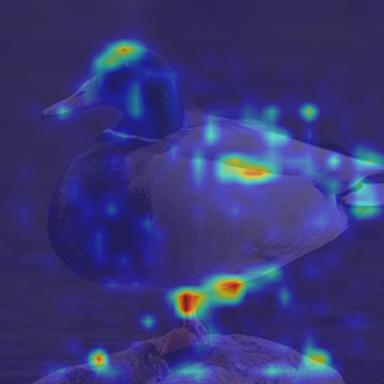}
            \caption*{\scriptsize LLaVA-OV-7b\\ \textit{\textcolor{red}{no}}}
        \end{minipage}
        \begin{minipage}[t]{0.13\textwidth}
            \includegraphics[width=\textwidth]{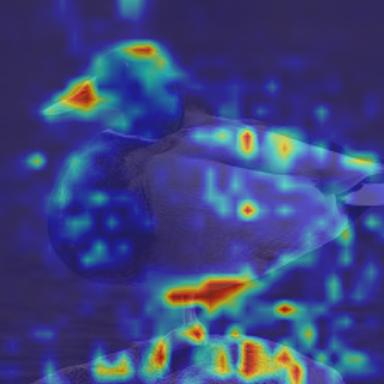}
            \caption*{\scriptsize LLaVA-OV-72b\\ \textit{No, the duck is not floating; it is standing on a \textcolor{red}{rock}.}}
        \end{minipage}
        \begin{minipage}[t]{0.13\textwidth}
            \includegraphics[width=\textwidth]{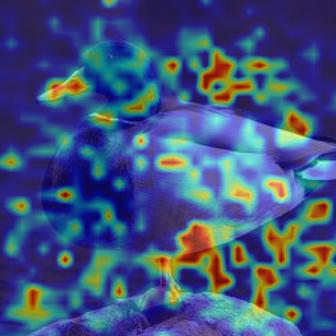}
            \caption*{\scriptsize Cambrian-3b\\ \textit{No, the duck is \textcolor{red}{standing} on a \textcolor{red}{rock}.}}
        \end{minipage}
        \begin{minipage}[t]{0.13\textwidth}
            \includegraphics[width=\textwidth]{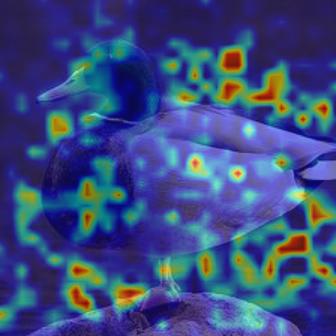}
            \caption*{\scriptsize Cambrian-8b\\ \textit{\textcolor{red}{No}}}
        \end{minipage}
        \begin{minipage}[t]{0.13\textwidth}
            \includegraphics[width=\textwidth]{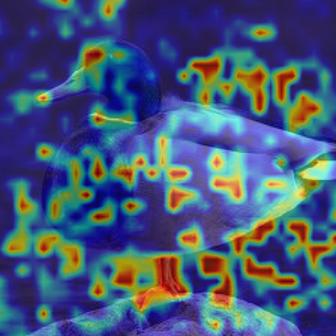}
            \caption*{\scriptsize Cambrian-13b\\ \textit{No, the duck is not floating. It is \textcolor{red}{standing} on a \textcolor{red}{rock}, which is \textcolor{red}{evident} from the \textcolor{red}{view} of its legs and feet.}}
        \end{minipage}
            \vspace{-0.4cm}
    \end{subfigure}

        \begin{subfigure}{0.98\textwidth}
        \centering
        \caption*{Question: Is the elderly person in the picture standing or sitting?}

        \begin{minipage}[t]{0.13\textwidth}
            \centering
            \includegraphics[width=\textwidth]{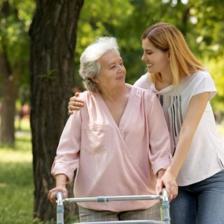}
            \caption*{\scriptsize Input image}
        \end{minipage}
        \begin{minipage}[t]{0.13\textwidth}
            \centering
            \includegraphics[width=\textwidth]{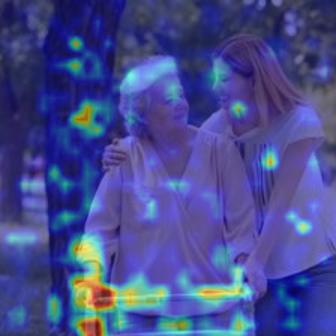}
            \caption*{\scriptsize LLaVA-1.5-7b\\ \textit{The elderly person in the picture is sitting \textcolor{red}{in} a \textcolor{red}{wheel}char.}}
        \end{minipage}
        \begin{minipage}[t]{0.13\textwidth}
            \centering
            \includegraphics[width=\textwidth]{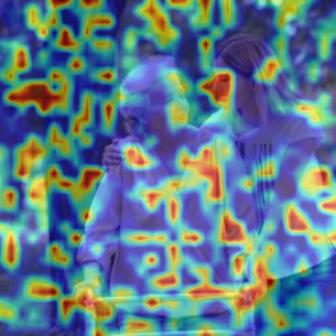}
            \caption*{\scriptsize Mini-Gemini-7b\\ \textit{The elderly person in the picture is sitting \textcolor{red}{in} a \textcolor{red}{walk}er.}}
        \end{minipage}
        \begin{minipage}[t]{0.13\textwidth}
            \centering
            \includegraphics[width=\textwidth]{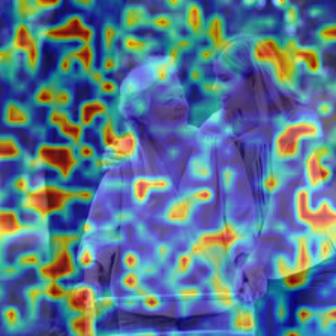}
            \caption*{\scriptsize Mini-Gemini-7b-HD\\ \textit{The elderly person in the picture is sitting in a \textcolor{red}{wheel}chair.}}
        \end{minipage}
        \begin{minipage}[t]{0.13\textwidth}
            \centering
            \includegraphics[width=\textwidth]{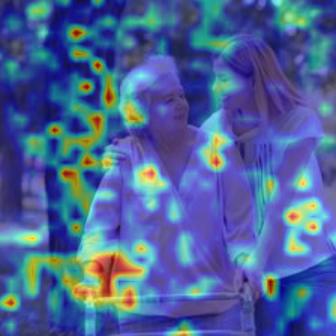}
            \caption*{\scriptsize Cambrian-13b\\ \textit{The elderly person in the picture is sitting \textcolor{red}{in} a \textcolor{red}{wheelchair}.}}
        \end{minipage}
        \begin{minipage}[t]{0.13\textwidth}
            \centering
            \includegraphics[width=\textwidth]{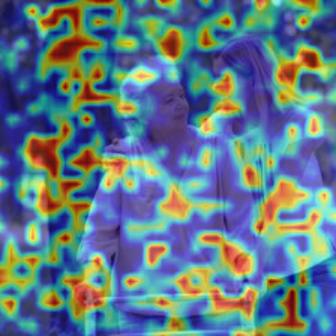}
            \caption*{\scriptsize Mini-Gemini-13b\\ \textit{The elderly person in the picture is sitting \textcolor{red}{in} a \textcolor{red}{walk}er.}}
        \end{minipage}
        \begin{minipage}[t]{0.13\textwidth}
            \centering
            \includegraphics[width=\textwidth]{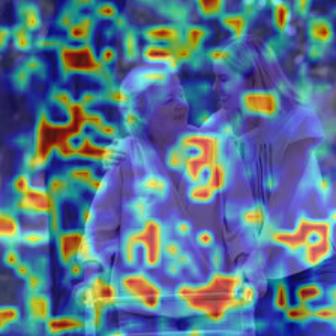}
            \caption*{\scriptsize Mini-Gemini-13b-HD\\ \textit{The elderly person in the picture is sitting \textcolor{red}{in} a \textcolor{red}{wheel}chair.}}
        \end{minipage}
    \end{subfigure}
    \captionsetup{justification=justified}
    \vspace{-0.2cm}
    \caption{Comparison of models with different LLM scales and vision architectures. The first row compares models with varying LLM sizes while keeping the vision architecture fixed, showing that increasing the LLM scale has minimal impact on visual behavior. The second row compares models with the same LLM but different vision encoders, indicating vision architectures may affect the focus region.}
    \vspace{-0.3cm}
    \label{fig:modelsize}
\end{figure*}

\noindent\textbf{Influence of Vision Architecture and LLM.}
Since the LVLMs consist of vision encoders and LLM, we investigate the impact of them on the focus region, respectively. To study the influence of LLM scale, we compare LLaVA-OV 0.5b, 7b, 72b and Cambrian 3b, 8b, 13b. For models with the same LLM but different vision architectures, we further include Mini-Gemini~\cite{li2024mini} as they provide high-resolution models (denoted as HD) with an additional vision encoder. The quantitative results are shown in Table~\ref{tab:scale} and we show qualitative results in Figure~\ref{fig:modelsize}.
It can be observed that merely increasing the LLM scale does not essentially change the focus region, despite the differences in response phrasing.
In contrast, given the same LLM, varying the vision architecture more significantly affects the focus regions. To verify these observations, we conduct a statistical significance test. It shows a significant impact of vision architecture (p=0.0008) but no statistical significance for LLM scale (p=0.121), which aligns with our observations. We show more visualization results in Appendix~\ref{sec:app_scale}.

%% file: 5_conclusion.tex
\vspace{-0.2cm}
\section{Conclusion}
\label{sec:conclusion}
\vspace{-0.1cm}
In this paper we propose a method to generalize existing visual interpretation methods to support the autoregressive, open-ended responses of LVLMs. We introduce a visually relevant token selection strategy that detects the crucial tokens in variable-length outputs and associates them with specific image regions. With the interpretation method, we conduct a comprehensive analysis of state-of-the-art open-source LVLMs with diverse model structures on visual instruction following benchmarks that require visual information. The experiment results provide several insights into model behaviors, including the relevance of the responses to visual input, relationship between answer correctness and focus region, influence of vision architectures and LLM scales. Despite some limitations discussed in Appendix~\ref{sec:app_limitation}, these findings emphasize the need for evaluation beyond standard accuracy metrics, offering insights into potential improvements of LVLMs.

%% file: 6_appendix.tex
\clearpage
\setcounter{page}{1}
\maketitlesupplementary

\noindent For complementary we conduct additional ablation studies on the parameter settings. We also compare the visualized model focus region with human attention. Besides, more qualitative results of the responses and the corresponding focus regions are presented to compare the visual behaviors of different model structures. Finally, we include a short discussion about the limitations of this work.

\section{Additional Ablation Studies}
\label{sec:app_ablation}
For our proposed heatmap visualization method, there is an important technical improvement to achieve better optimization, which introduces graduated non-convexity by adding an exponentially decayed $L2$ norm. In this section, we conduct parameter studies to evaluate the impact of the weight of the $L2$ norm. We use LLaVA-v1.5-7b for the following experiments. We first vary the scale of the $L2$ norm $\lambda_2$ among $\left[0.0, 0.1, 1.0, 10.0\right]$ and compare the deletion and insertion scores on different datasets, as shown in Table~\ref{sup:lambda}. It demonstrates that introducing a proper scale of $L2$ norm can benefit the optimization and obtain better results in terms of the deletion and insertion. However, if the scale of the regularization is too large, it may disturb the objective function and degrade the performance. We select $\lambda_2=0.1$ and $\alpha=0.2$ for the exponential decay rate.

\begin{table}[h]
\centering
\begin{tabular}{lcccccc}
\hline
                 & \multicolumn{2}{c}{MMVP}        & \multicolumn{2}{c}{MMStar}      & \multicolumn{2}{c}{CV-Bench}    \\ \hline
 $\lambda_2$                & Del$\downarrow$ & Ins$\uparrow$ & Del$\downarrow$ & Ins$\uparrow$ & Del$\downarrow$ & Ins$\uparrow$ \\ \hline
0.0    & 0.381           & 0.794         & 0.295           & \textbf{0.953}         & \textbf{0.398}           & \textbf{0.965}         \\
0.1  & \textbf{0.366}           & \textbf{0.811}         & \textbf{0.292}           & \textbf{0.953}         & 0.402           & \textbf{0.965}         \\
1.0  & 0.514           & 0.705         & 0.420           & 0.846         & 0.524           & 0.885         \\
10.0 & 0.513           & 0.735         & 0.412           & 0.858         & 0.520           & 0.894         \\ \hline
\end{tabular}
\caption{Parameter study of the scale of the $L2$ norm in graduated non-convexity. The bolded numbers denote the best results among the compared parameter settings. In the main experiments we select $\lambda_2=0.1$. }
\label{sup:lambda}
\end{table}

\section{Comparison with Human Attention}
\label{sec:app_human}
In this paper we mainly investigate the intrinsic behavior of the LVLMs in terms of the focus region when generating the open-ended responses. The model attention is an objective fact which does not need to be aligned with human attention/behavior. However, we conduct experiments to compare the model behavior with human attention to gain deeper insights. We evaluate the LVLMs on VQA-HAT dataset~\cite{vqahat}, which consists of human visual attention maps over the images in the VQA dataset~\cite{antol2015vqa}. In Table~\ref{humanatt} we show the soft IOU and rank correlation between the focus region of LLaVA-1.5/LLaVA-OV and human attention labels on VQA-HAT dataset. IOU measures the overlap between the model attention maps and human attention maps, defined as the ratio of their intersection to their union and the range is between 0 and 1. Rank correlation is a statistical measure that quantifies the similarity between the rankings of two variables, commonly used to assess monotonic relationships. The results indicate that the focus regions of LVLMs can significantly differ from human attention, with small IOU and negative rank correlation values.

\begin{table}[]
\centering
\begin{tabular}{lcc}
\hline
 & IOU & Rank correlation \\ \hline
LLaVA-1.5 & 0.010 & -0.201$\pm$0.003\\
LLaVA-OV & 0.012 &  -0.195$\pm$0.004\\ \hline
\end{tabular}
\caption{Comparison of model focus region with human attention on VQA dataset, evaluated by the IOU and rank correlation.}
\label{humanatt}
\end{table}

\section{Influence of the Vision Architecture and LLM scale}
\label{sec:app_scale}
\begin{figure*}[tbp]
    \captionsetup{justification=centering}
    \centering
    \captionsetup{font=small}
    \begin{subfigure}{0.49\textwidth}
        \centering
        \caption*{Question: From which angle is this image taken?}
        \begin{minipage}[t]{0.24\textwidth}
            \centering
            \includegraphics[width=\textwidth]{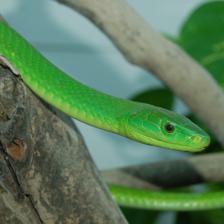}
            \caption*{\scriptsize Input image}
        \end{minipage}
        \begin{minipage}[t]{0.24\textwidth}
            \centering
            \includegraphics[width=\textwidth]{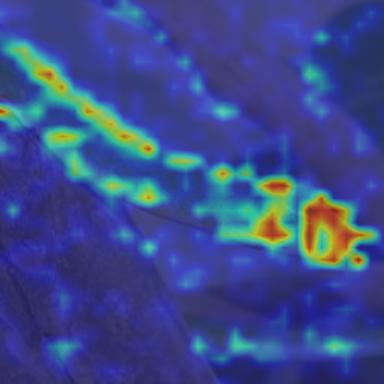}
            \caption*{\scriptsize LLaVA-OV-0.5b\\ \textit{The image is taken from a \textcolor{red}{side} angle, \textcolor{red}{showing} the \textcolor{red}{snake}'s \textcolor{red}{profile}.}}
        \end{minipage}
        \begin{minipage}[t]{0.24\textwidth}
            \centering
            \includegraphics[width=\textwidth]{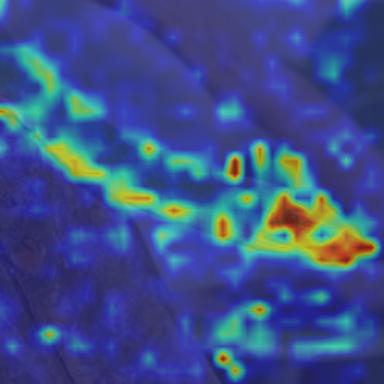}
            \caption*{\scriptsize LLaVA-OV-7b\\ \textit{The image is taken from a \textcolor{red}{side} angle, \textcolor{red}{slightly above} the \textcolor{red}{snake}. The \textcolor{red}{snake}'s body is \textcolor{red}{visible} in \textcolor{red}{profile}, and its head is \textcolor{red}{turned} towards the camera, giving a clear view of its \textcolor{red}{eye} and the pattern on its skin.}}
        \end{minipage}
        \begin{minipage}[t]{0.24\textwidth}
            \centering
            \includegraphics[width=\textwidth]{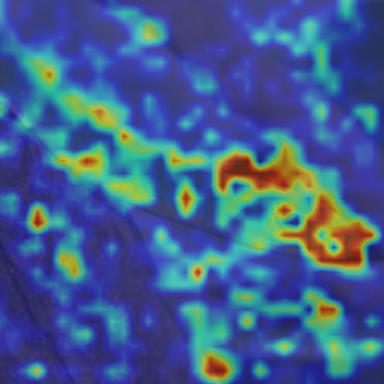}
            \caption*{\scriptsize LLaVA-OV-72b\\ \textit{The image is taken from a \textcolor{red}{side} angle, \textcolor{red}{slightly} above the \textcolor{red}{level} of the \textcolor{red}{snake}'s body. This perspective allows for a clear view of the snake's \textcolor{red}{profile} and the texture of its scales.}}
        \end{minipage}
    \end{subfigure}
    \begin{subfigure}{0.49\textwidth}
        \centering
        \caption*{Question: What does the center button say?}

        \begin{minipage}[t]{0.24\textwidth}
            \centering
            \includegraphics[width=\textwidth]{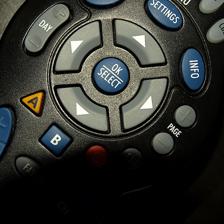}
            \caption*{\scriptsize Input image}
        \end{minipage}
        \begin{minipage}[t]{0.24\textwidth}
            \centering
            \includegraphics[width=\textwidth]{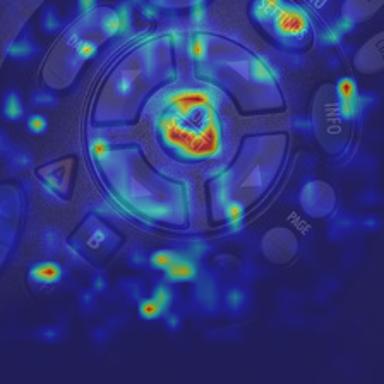}
            \caption*{\scriptsize LLaVA-OV-0.5b\\ \textit{The center button says "\textcolor{red}{OK SELECT}".}}
        \end{minipage}
        \begin{minipage}[t]{0.24\textwidth}
            \centering
            \includegraphics[width=\textwidth]{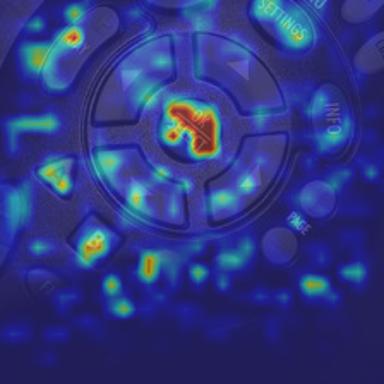}
            \caption*{\scriptsize LLaVA-OV-7b\\ \textit{The center button \textcolor{red}{on} the \textcolor{red}{remote} control \textcolor{red}{in} the image is labeled "\textcolor{red}{OK SELECT}."}}
        \end{minipage}
        \begin{minipage}[t]{0.24\textwidth}
            \centering
            \includegraphics[width=\textwidth]{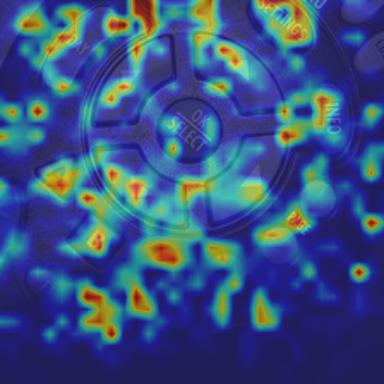}
            \caption*{\scriptsize LLaVA-OV-72b\\ \textit{The center button on the \textcolor{red}{remote} control is labeled 'OK' and \textcolor{red}{also} has the \textcolor{red}{word} '\textcolor{red}{SELECT}' \textcolor{red}{beneath} it, indicating that it is used to confirm selections or navigate through menus on a device such as a television or a \textcolor{red}{cable} box.}}
        \end{minipage}
    \end{subfigure}

    \begin{subfigure}{0.49\textwidth}
        \centering
        \caption*{Question: Is the statement correct: There is no letter D on this image.}

        \begin{minipage}[t]{0.24\textwidth}
            \centering
            \includegraphics[width=\textwidth]{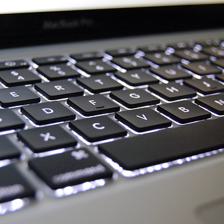}
            \caption*{\scriptsize Input image}
        \end{minipage}
        \begin{minipage}[t]{0.24\textwidth}
            \centering
            \includegraphics[width=\textwidth]{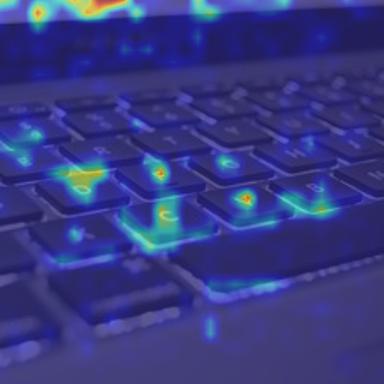}
            \caption*{\scriptsize LLaVA-OV-0.5b\\ \textit{No, the \textcolor{red}{letter} D is present on this image.}}
        \end{minipage}
        \begin{minipage}[t]{0.24\textwidth}
            \centering
            \includegraphics[width=\textwidth]{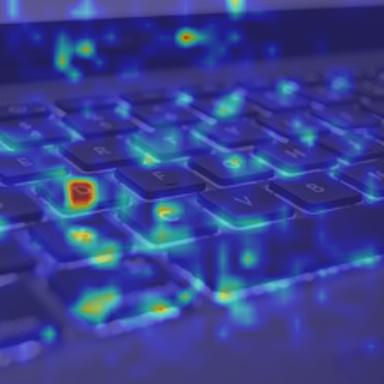}
            \caption*{\scriptsize LLaVA-OV-7b\\ \textit{The statement is \textcolor{red}{incorrect}. There is a letter "D" on the \textcolor{red}{keyboard} in the image, specifically in the \textcolor{red}{second} row from the top, towards the left side of the image.}}
        \end{minipage}
        \begin{minipage}[t]{0.24\textwidth}
            \centering
            \includegraphics[width=\textwidth]{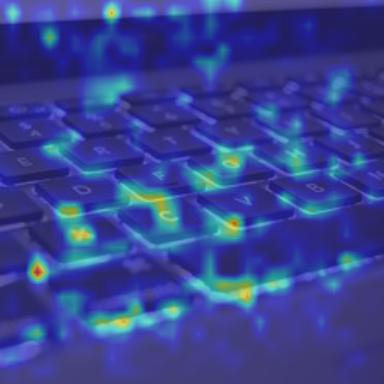}
            \caption*{\scriptsize LLaVA-OV-72b\\ \textit{No, the statement is incorrect. The \textcolor{red}{letter} D is \textcolor{red}{visible} on the \textcolor{red}{keyboard} in the image.}}
        \end{minipage}
    \end{subfigure}
    \begin{subfigure}{0.49\textwidth}
        \centering
        \caption*{Question: Are there any clouds?}

        \begin{minipage}[t]{0.24\textwidth}
            \centering
            \includegraphics[width=\textwidth]{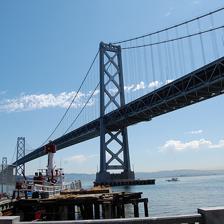}
            \caption*{\scriptsize Input image}
        \end{minipage}
        \begin{minipage}[t]{0.24\textwidth}
            \centering
            \includegraphics[width=\textwidth]{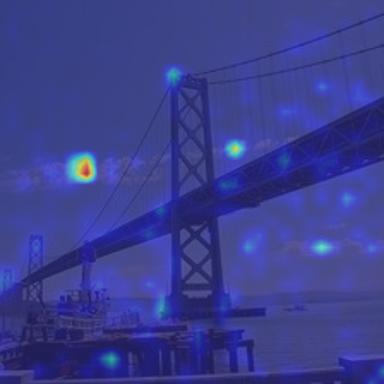}
            \caption*{\scriptsize LLaVA-OV-0.5b\\ \textit{Yes, there are some \textcolor{red}{scattered} clouds in the sky.}}
        \end{minipage}
        \begin{minipage}[t]{0.24\textwidth}
            \centering
            \includegraphics[width=\textwidth]{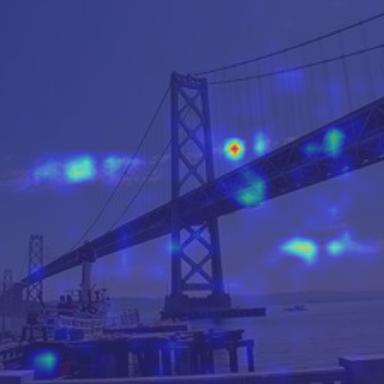}
            \caption*{\scriptsize LLaVA-OV-7b\\ \textit{\textcolor{red}{yes}}}
        \end{minipage}
        \begin{minipage}[t]{0.24\textwidth}
            \centering
            \includegraphics[width=\textwidth]{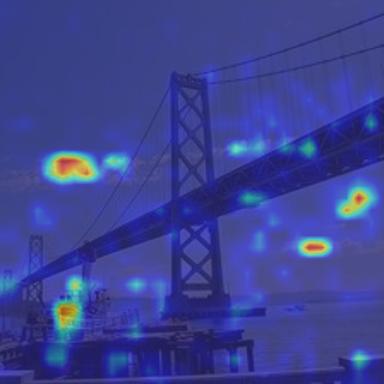}
            \caption*{\scriptsize LLaVA-OV-72b\\ \textit{Yes, there are \textcolor{red}{a} few scattered clouds in the sky.}}
        \end{minipage}
    \end{subfigure}
    \caption{Comparison of the responses and focus region of LLaVA-OV with different LLM scales. The tokens in red denote the selected crucial tokens. The responses of the models with different scales often have different expressions, but the corresponding focus regions are often similar.}
    \label{sup:fig_scale1}
\end{figure*}

\begin{figure*}[tbp]
    \captionsetup{justification=centering}
    \centering
    \captionsetup{font=small}
        \begin{subfigure}{0.49\textwidth}
        \centering
        \caption*{Question: Is the shark's belly visible in this image?}

        \begin{minipage}[t]{0.24\textwidth}
            \centering
            \includegraphics[width=\textwidth]{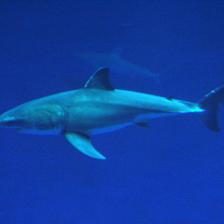}
            \caption*{\scriptsize Input image}
        \end{minipage}
        \begin{minipage}[t]{0.24\textwidth}
            \centering
            \includegraphics[width=\textwidth]{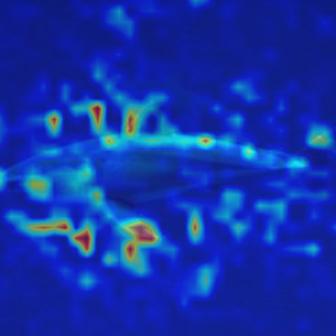}
            \caption*{\scriptsize Cambrian-3b\\ \textit{No, the shark's belly is not visible in this image. The \textcolor{red}{angle} and the position of the shark, combined with the lighting and the \textcolor{red}{depth} of the water, obscure the view of its underside.}}
        \end{minipage}
        \begin{minipage}[t]{0.24\textwidth}
            \centering
            \includegraphics[width=\textwidth]{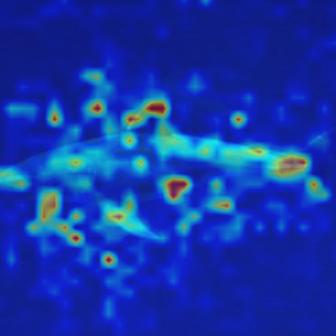}
            \caption*{\scriptsize Cambrian-8b\\ \textit{Yes, the shark's belly is visible in this image. It appears to be a \textcolor{red}{lighter} shade of blue compared to the rest of its body, which is \textcolor{red}{typical} for many shark species as their underside is often less dark than their dorsal side.}}
        \end{minipage}
        \begin{minipage}[t]{0.24\textwidth}
            \centering
            \includegraphics[width=\textwidth]{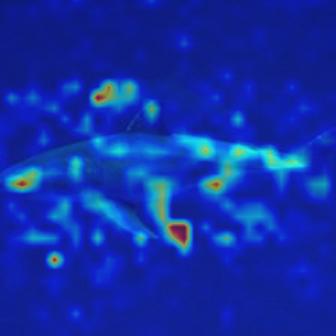}
            \caption*{\scriptsize Cambrian-13b\\ \textit{Yes, the shark's belly is visible in this image. It appears to be a \textcolor{red}{l}ighter shade of blue compared to the rest of its body, which is typical for many shark species as it helps them blend in with the ocean's surface when viewed from below.}}
        \end{minipage}
    \end{subfigure}
    \hfill
    \begin{subfigure}{0.49\textwidth}
        \centering
        \caption*{Question: Can you see letter "J" in the image?}

        \begin{minipage}[t]{0.24\textwidth}
            \centering
            \includegraphics[width=\textwidth]{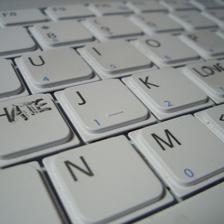}
            \caption*{\scriptsize Input image}
        \end{minipage}
        \begin{minipage}[t]{0.24\textwidth}
            \centering
            \includegraphics[width=\textwidth]{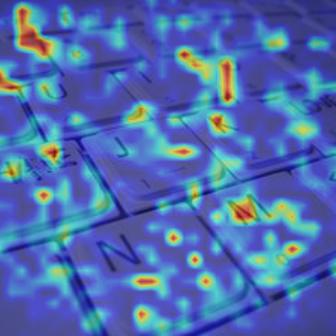}
            \caption*{\scriptsize Cambrian-3b\\ \textit{Yes, the letter 'J' is clearly visible \textcolor{red}{on} the \textcolor{red}{keyboard}, located in the second row from the top and the third column from the left.}}
        \end{minipage}
        \begin{minipage}[t]{0.24\textwidth}
            \centering
            \includegraphics[width=\textwidth]{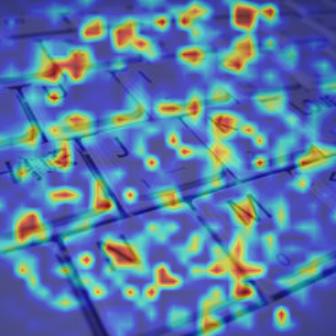}
            \caption*{\scriptsize Cambrian-8b\\ \textit{Yes, the letter "J" is clearly visible in the image, \textcolor{red}{located on} the \textcolor{red}{keyboard}.}}
        \end{minipage}
        \begin{minipage}[t]{0.24\textwidth}
            \centering
            \includegraphics[width=\textwidth]{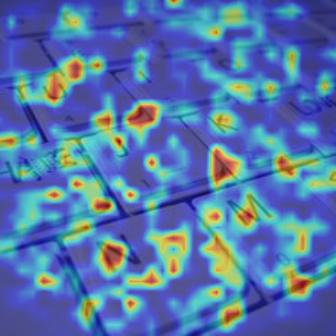}
            \caption*{\scriptsize Cambrian-13b\\ \textit{Yes, the letter "J" is clearly visible \textcolor{red}{on} the \textcolor{red}{keyboard}. It is on the top row, second from the left. The "J" key is \textcolor{red}{white} with black lettering, which is a standard color scheme for keys on many keyboards. The \textcolor{red}{key} is in focus and there are no obstructions.}}
        \end{minipage}
    \end{subfigure}
    
    \begin{subfigure}{0.49\textwidth}
        \centering
        \caption*{Question: How many spots are on the animal in the image?}
        \begin{minipage}[t]{0.24\textwidth}
            \centering
            \includegraphics[width=\textwidth]{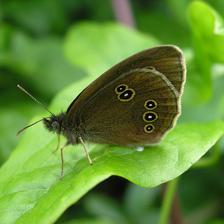}
            \caption*{\scriptsize Input image}
        \end{minipage}
        \begin{minipage}[t]{0.24\textwidth}
            \centering
            \includegraphics[width=\textwidth]{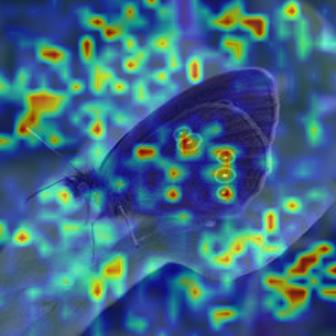}
            \caption*{\scriptsize Cambrian-3b\\ \textit{There are \textcolor{red}{three} spots on the animal in the image.}}
        \end{minipage}
        \begin{minipage}[t]{0.24\textwidth}
            \centering
            \includegraphics[width=\textwidth]{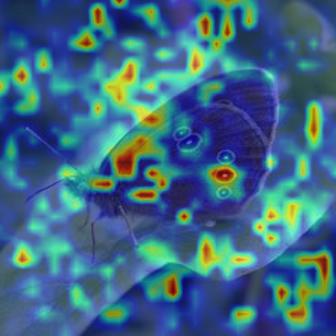}
            \caption*{\scriptsize Cambrian-8b\\ \textit{There \textcolor{red}{are three} spots on the animal in the image.}}
        \end{minipage}
        \begin{minipage}[t]{0.24\textwidth}
            \centering
            \includegraphics[width=\textwidth]{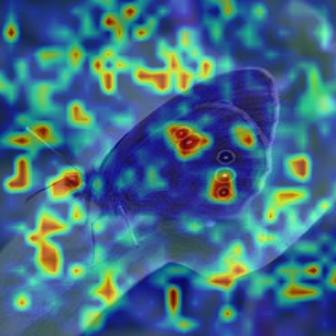}
            \caption*{\scriptsize Cambrian-13b\\ \textit{There are \textcolor{red}{three} spots on the animal in the image.}}
        \end{minipage}
    \end{subfigure}
    \begin{subfigure}{0.49\textwidth}
        \centering
        \caption*{Question: How many cup(s) of drink are there in the image?}

        \begin{minipage}[t]{0.24\textwidth}
            \centering
            \includegraphics[width=\textwidth]{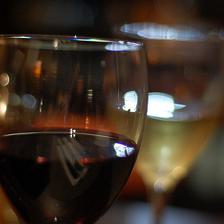}
            \caption*{\scriptsize Input image}
        \end{minipage}
        \begin{minipage}[t]{0.24\textwidth}
            \centering
            \includegraphics[width=\textwidth]{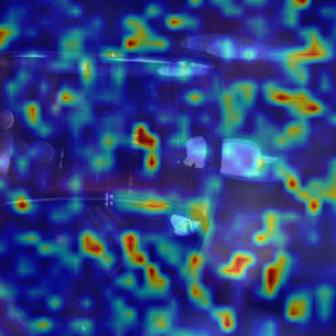}
            \caption*{\scriptsize Cambrian-3b\\ \textit{There are \textcolor{red}{two} cup(s) of drink in the image.}}
        \end{minipage}
        \begin{minipage}[t]{0.24\textwidth}
            \centering
            \includegraphics[width=\textwidth]{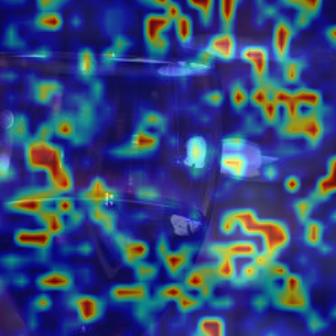}
            \caption*{\scriptsize Cambrian-8b\\ \textit{\textcolor{red}{2}}}
        \end{minipage}
        \begin{minipage}[t]{0.24\textwidth}
            \centering
            \includegraphics[width=\textwidth]{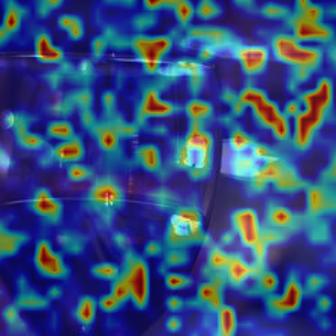}
            \caption*{\scriptsize Cambrian-13b\\ \textit{\textcolor{red}{2}}}
        \end{minipage}
    \end{subfigure}
    \caption{Comparison of the responses and focus region of Cambrian with different LLM scales. The tokens in red denote the selected crucial tokens. Cambrian often tend to attend to the whole image for more comprehensive visual understanding.}
    \label{sup:fig_scale2}
\end{figure*}

\begin{figure*}[tbp]
    \captionsetup{justification=centering}
    \centering
    \captionsetup{font=small}    
    \begin{subfigure}{\textwidth}
        \centering
        \caption*{Question: How many spots are on the animal in the image?}

        \begin{minipage}[t]{0.135\textwidth}
            \centering
            \includegraphics[width=\textwidth]{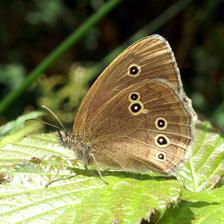}
            \caption*{\scriptsize Input image}
        \end{minipage}
        \begin{minipage}[t]{0.135\textwidth}
            \centering
            \includegraphics[width=\textwidth]{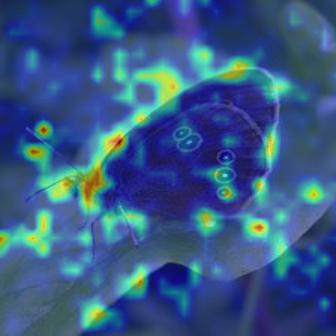}
            \caption*{\scriptsize LLaVA-1.5-7b\\ \textit{The animal in the image has many spots on its \textcolor{red}{wings}.}}
        \end{minipage}
        \begin{minipage}[t]{0.135\textwidth}
            \centering
            \includegraphics[width=\textwidth]{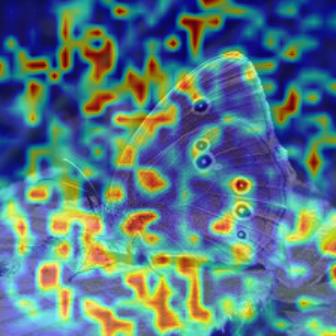}
            \caption*{\scriptsize Mini-Gemini-7b\\ \textit{The image \textcolor{red}{shows} a \textcolor{red}{but}terfly with many spots on its wings.}}
        \end{minipage}
        \begin{minipage}[t]{0.135\textwidth}
            \centering
            \includegraphics[width=\textwidth]{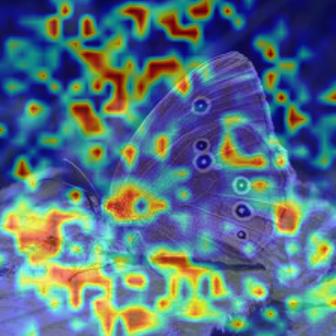}
            \caption*{\scriptsize Mini-Gemini-7b-HD\\ \textit{The image shows a \textcolor{red}{but}terfly with many spots on its wings. However, I cannot provide an exact number of spots without more specific information about the butterfly species.}}
        \end{minipage}
        \begin{minipage}[t]{0.135\textwidth}
            \centering
            \includegraphics[width=\textwidth]{14_cambrian13b.jpg}
            \caption*{\scriptsize Cambrian-13b\\ \textit{There \textcolor{red}{are three} spots on the animal in the image.}}
        \end{minipage}
        \begin{minipage}[t]{0.135\textwidth}
            \centering
            \includegraphics[width=\textwidth]{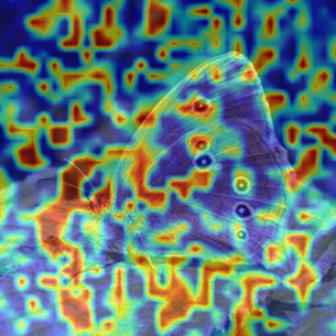}
            \caption*{\scriptsize Mini-Gemini-13b\\ \textit{The \textcolor{red}{animal} in the image, which is a \textcolor{red}{but}terfly, has \textcolor{red}{three} spots on its wings.}}
        \end{minipage}
        \begin{minipage}[t]{0.135\textwidth}
            \centering
            \includegraphics[width=\textwidth]{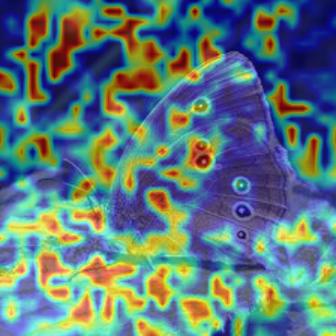}
            \caption*{\scriptsize Mini-Gemini-13b-HD\\ \textit{The \textcolor{red}{animal} in the image, a \textcolor{red}{but}terfly, has \textcolor{red}{two} spots on its wings.}}
        \end{minipage}
    \end{subfigure}
    \begin{subfigure}{\textwidth}
        \centering
        \caption*{Question: Where is the yellow animal's head lying in this image?}

        \begin{minipage}[t]{0.135\textwidth}
            \centering
            \includegraphics[width=\textwidth]{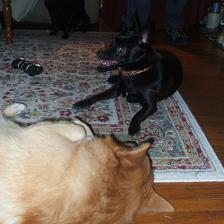}
            \caption*{\scriptsize Input image}
        \end{minipage}
        \begin{minipage}[t]{0.135\textwidth}
            \centering
            \includegraphics[width=\textwidth]{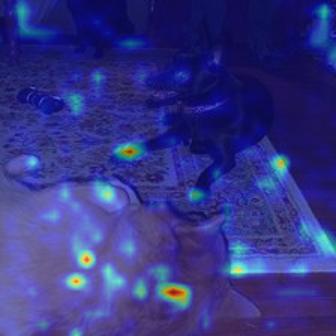}
            \caption*{\scriptsize LLaVA-1.5-7b\\ \textit{The yellow animal's head is lying on the \textcolor{red}{floor} in this image.}}
        \end{minipage}
        \begin{minipage}[t]{0.135\textwidth}
            \centering
            \includegraphics[width=\textwidth]{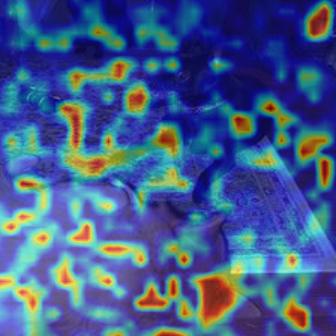}
            \caption*{\scriptsize Mini-Gemini-7b\\ \textit{The yellow animal's head is lying on the \textcolor{red}{floor} in this image.}}
        \end{minipage}
        \begin{minipage}[t]{0.135\textwidth}
            \centering
            \includegraphics[width=\textwidth]{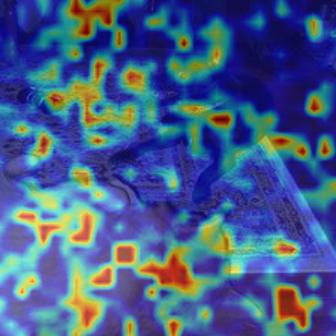}
            \caption*{\scriptsize Mini-Gemini-7b-HD\\ \textit{The yellow animal's head is lying on the \textcolor{red}{floor} in the image.}}
        \end{minipage}
        \begin{minipage}[t]{0.135\textwidth}
            \centering
            \includegraphics[width=\textwidth]{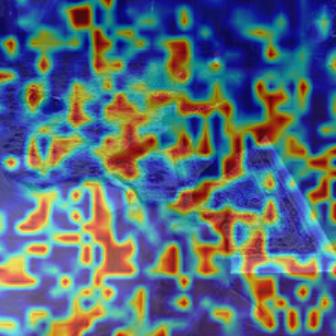}
            \caption*{\scriptsize Cambrian-13b\\ \textit{The yellow animal's head is lying on the \textcolor{red}{floor}, \textcolor{red}{resting} comfortably on the \textcolor{red}{patterned} rug.}}
        \end{minipage}
        \begin{minipage}[t]{0.135\textwidth}
            \centering
            \includegraphics[width=\textwidth]{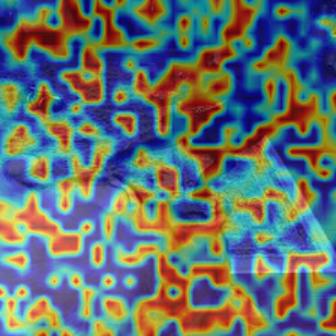}
            \caption*{\scriptsize Mini-Gemini-13b\\ \textit{The yellow animal's head is lying on the \textcolor{red}{floor}, next to the \textcolor{red}{black dog}.}}
        \end{minipage}
        \begin{minipage}[t]{0.135\textwidth}
            \centering
            \includegraphics[width=\textwidth]{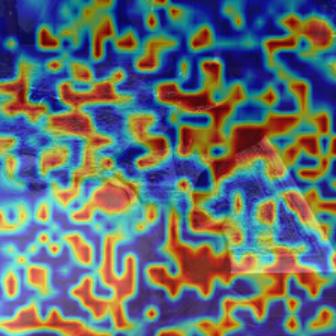}
            \caption*{\scriptsize Mini-Gemini-13b-HD\\ \textit{The yellow animal's head is lying on the \textcolor{red}{floor}, next to a \textcolor{red}{black dog}.}}
        \end{minipage}
    \end{subfigure}

    \captionsetup{justification=justified}
    \caption{Comparison of the focus region of LVLMs with different vision architectures and LLM scales. The tokens in red denote the selected crucial tokens. \emph{HD} denotes the high-resolution vision encoder.}
    \label{sup:fig_scale3}
\end{figure*}

In the main experiments, we studied the influence of the vision architecture and LLM scale on the visual behavior (\emph{i.e.,} focus region when generating the responses). Here we present more qualitative results to compare the LVLMs with different vision architectures and LLM scales, respectively. In Figure~\ref{sup:fig_scale1}, we compare the responses and corresponding focus region of LLaVA-OV 0.5b, 7b, 72b. In Figure~\ref{sup:fig_scale2}, we compare the responses and corresponding focus region of Cambrian 3b, 8b, 13b. In Figure~\ref{sup:fig_scale3}, we compare LLaVA-1.5-7b, Mini-Gemini-7b, and Mini-Gemini-7b-HD, as well as Cambrian-13b with Mini-Gemini-13b, and Mini-Gemini-13b-HD. Within each group of models, they use the same LLM (\emph{i.e.,} Vicuna1.5-7b and Vicuna1.5-13b, respectively). It can be observed that the responses of the models with different LLM scales may have different expressions, but the corresponding focus regions often have similar structures. It indicates that the focus region is not significantly affected by the scale of the LLM. Instead, the LLM scale primarily influences the linguistic expression of the responses. In contrast, with the same LLM, vision architecture may have more significant influence on the focus region related to the model outputs.

\section{Additional Qualitative Results}
\label{app_visualize}
\begin{figure*}[htbp]
    \captionsetup{justification=centering}
    \centering
    \captionsetup{font=small}
    \begin{subfigure}{0.49\textwidth}
        \centering
        \caption*{Question: Is the flower's center protruding or indented?}
        \begin{minipage}[t]{0.24\textwidth}
            \includegraphics[width=\textwidth]{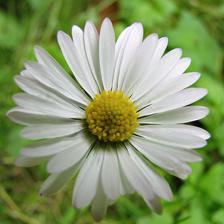}
            \caption*{\scriptsize Input image}
        \end{minipage}
        \hfill
        \begin{minipage}[t]{0.24\textwidth}
            \includegraphics[width=\textwidth]{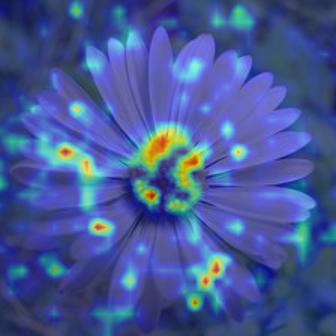}
            \caption*{\scriptsize LLaVA-1.5\\ \textit{The flower's center is protruding, as it has a \textcolor{red}{yellow} center and a white \textcolor{red}{pet}al.}}
        \end{minipage}
        \hfill
        \begin{minipage}[t]{0.24\textwidth}
            \includegraphics[width=\textwidth]{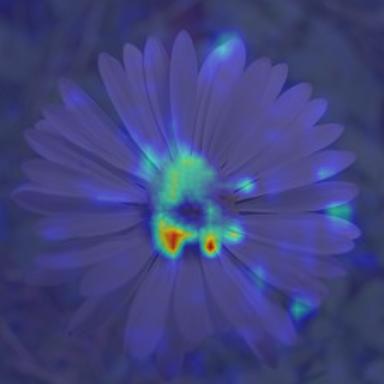}
            \caption*{\scriptsize LLaVA-OV \\ \textit{The flower's center is protruding, as indicated by the text "The flower's center is \textcolor{red}{protr}uding."}}
        \end{minipage}
        \hfill
        \begin{minipage}[t]{0.24\textwidth}
            \includegraphics[width=\textwidth]{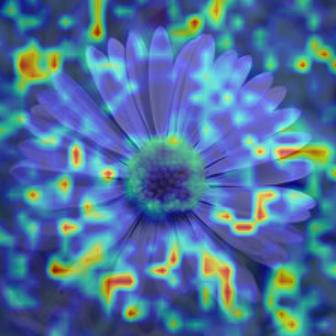}
            \caption*{\scriptsize Cambrian \\ \textit{The flower's center is not \textcolor{red}{protr}uding but rather appears to be indented. This is indicated by the way the petals are arranged around the center, with the edges of the petals slightly overlapping the center, suggesting that the center is recessed into the petals.}}
        \end{minipage}
        
    \end{subfigure}
    \hfill
    \begin{subfigure}{0.49\textwidth}
        \centering
        \caption*{Question: Is the person weighing themselves barefoot?}
        \begin{minipage}[t]{0.24\textwidth}
            \includegraphics[width=\textwidth]{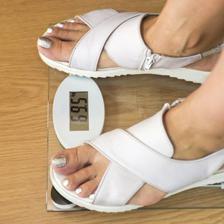}
            \caption*{\scriptsize Input image}
        \end{minipage}
        \hfill
        \begin{minipage}[t]{0.24\textwidth}
            \includegraphics[width=\textwidth]{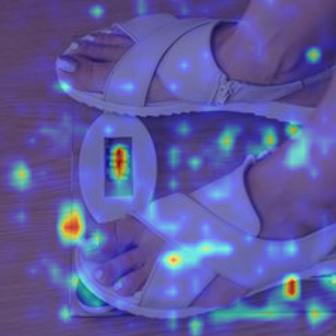}
            \caption*{\scriptsize LLaVA-1.5\\ \textit{Yes, the person is weighing themselves barefoot \textcolor{red}{on} the scale.}}
        \end{minipage}
        \hfill
        \begin{minipage}[t]{0.24\textwidth}
            \includegraphics[width=\textwidth]{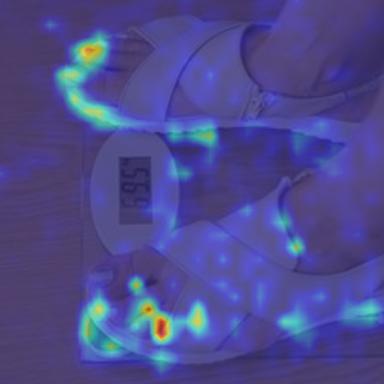}
            \caption*{\scriptsize LLaVA-OV \\ \textit{\textcolor{red}{no}.}}
        \end{minipage}
        \hfill
        \begin{minipage}[t]{0.24\textwidth}
            \includegraphics[width=\textwidth]{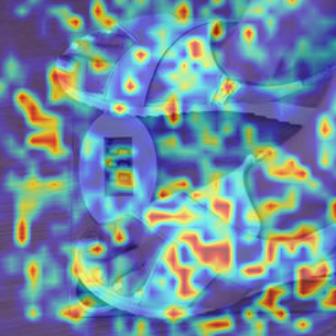}
            \caption*{\scriptsize Cambrian \\ \textit{Based on the image, the person is \textcolor{red}{not weighing} themselves barefoot. They are wearing \textcolor{red}{white sandals} with straps that are \textcolor{red}{fast}ened around the ankles.}}
        \end{minipage}
    \end{subfigure}
    
    \begin{subfigure}{0.49\textwidth}
        \centering
        \caption*{Question: Is there a pomegranate in the picture?}
        \begin{minipage}[t]{0.24\textwidth}
            \includegraphics[width=\textwidth]{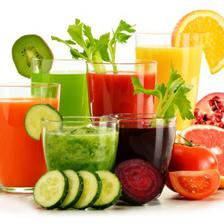}
            \caption*{\scriptsize Input image}
        \end{minipage}
        \hfill
        \begin{minipage}[t]{0.24\textwidth}
            \includegraphics[width=\textwidth]{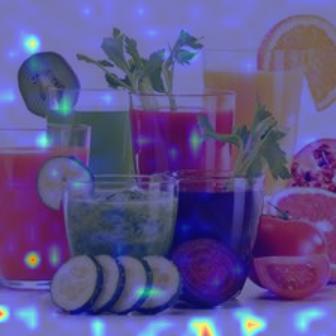}
            \caption*{\scriptsize LLaVA-1.5\\ \textit{Yes, there is a pomegranate in the picture, \textcolor{red}{along} with other fruits and vegetables.}}
        \end{minipage}
        \hfill
        \begin{minipage}[t]{0.24\textwidth}
            \includegraphics[width=\textwidth]{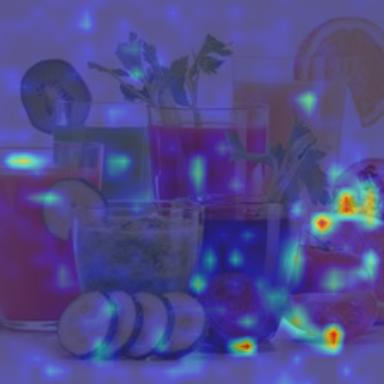}
            \caption*{\scriptsize LLaVA-OV \\ \textit{Yes, there is a pomegranate in the picture. \textcolor{red}{It}'s located on the \textcolor{red}{right} side of the image, near the top.}}
        \end{minipage}
        \hfill
        \begin{minipage}[t]{0.24\textwidth}
            \includegraphics[width=\textwidth]{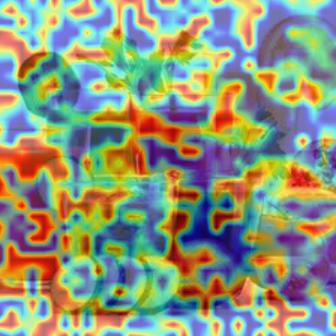}
            \caption*{\scriptsize Cambrian \\ \textit{Yes, there is a pomegranate in the picture. It is located on the right side of the image, \textcolor{red}{next} to the \textcolor{red}{glass} of \textcolor{red}{orange} juice. The pomegranate is \textcolor{red}{cut} open, revealing its juicy red seeds, which are a \textcolor{red}{common ingredient} in fruit \textcolor{red}{juices} and \textcolor{red}{smooth}ies.}}
        \end{minipage}
    \end{subfigure}
    \hfill
    \begin{subfigure}{0.49\textwidth}
        \centering
        \caption*{Question: Is the sky in the background dark blue or light blue?}
        \begin{minipage}[t]{0.24\textwidth}
            \includegraphics[width=\textwidth]{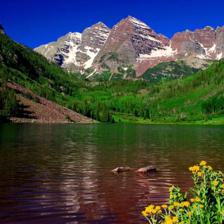}
            \caption*{\scriptsize Input image}
        \end{minipage}
        \hfill
        \begin{minipage}[t]{0.24\textwidth}
            \includegraphics[width=\textwidth]{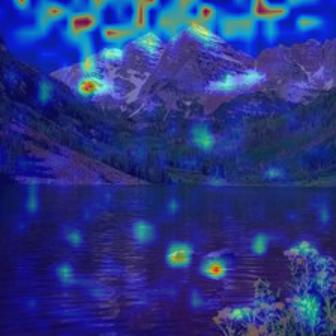}
            \caption*{\scriptsize LLaVA-1.5\\ \textit{The sky in the background is \textcolor{red}{light} blue.}}
        \end{minipage}
        \hfill
        \begin{minipage}[t]{0.24\textwidth}
            \includegraphics[width=\textwidth]{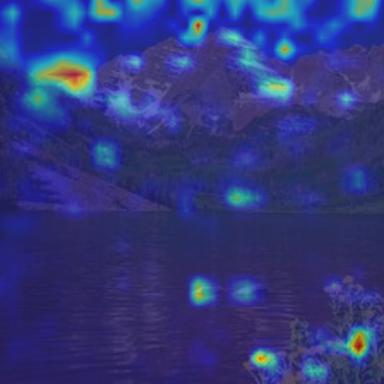}
            \caption*{\scriptsize LLaVA-OV \\ \textit{The sky in the background is a \textcolor{red}{light} blue color.}}
        \end{minipage}
        \hfill
        \begin{minipage}[t]{0.24\textwidth}
            \includegraphics[width=\textwidth]{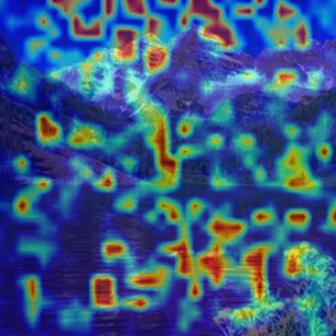}
            \caption*{\scriptsize Cambrian \\ \textit{The sky in the background is \textcolor{red}{light} blue, providing a serene backdrop to the \textcolor{red}{mountain}ous landscape.}}
        \end{minipage}
    \end{subfigure}

    \begin{subfigure}{0.49\textwidth}
        \centering
        \caption*{Question: How many balls are on the table?}
        \begin{minipage}[t]{0.24\textwidth}
            \includegraphics[width=\textwidth]{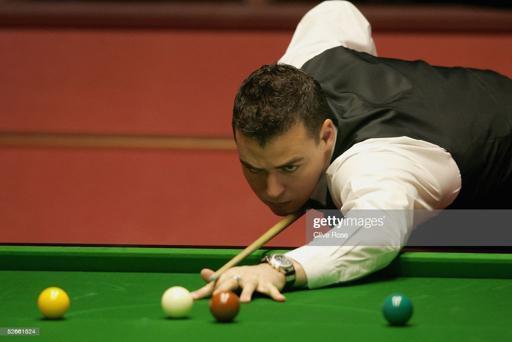}
            \caption*{\scriptsize Input image}
        \end{minipage}
        \begin{minipage}[t]{0.24\textwidth}
            \includegraphics[width=\textwidth]{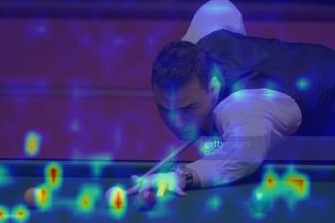}
            \caption*{\scriptsize LLaVA-1.5\\ \textit{There are \textcolor{red}{three} balls on the table.}}
        \end{minipage}
        \begin{minipage}[t]{0.24\textwidth}
            \includegraphics[width=\textwidth]{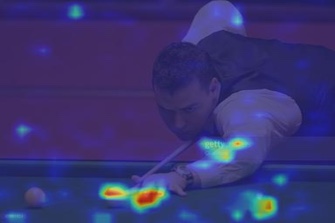}
            \caption*{\scriptsize LLaVA-OV \\ \textit{There are \textcolor{red}{four} balls on the table.}}
        \end{minipage}
        \begin{minipage}[t]{0.24\textwidth}
            \includegraphics[width=\textwidth]{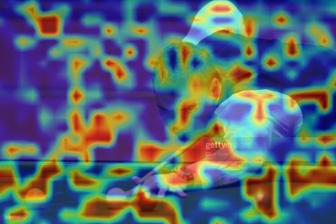}
            \caption*{\scriptsize Cambrian \\ \textit{There are \textcolor{red}{six} balls on the table, each with a distinct color and \textcolor{red}{number}, arranged in a \textcolor{red}{triangle} formation \textcolor{red}{at} the \textcolor{red}{start} of a game of \textcolor{red}{sn}ooker.}}
        \end{minipage}
    \end{subfigure}
    \hfill
    \begin{subfigure}{0.49\textwidth}
        \centering
        \caption*{Question: How many computer monitors are in this picture?}
        \begin{minipage}[t]{0.24\textwidth}
            \includegraphics[width=\textwidth]{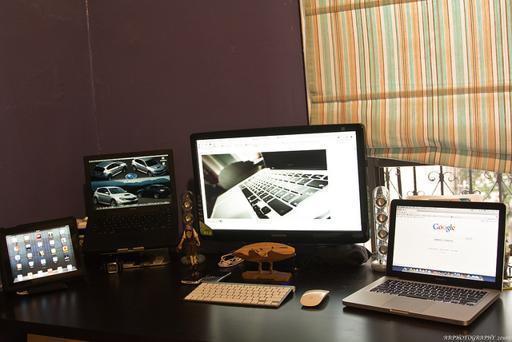}
            \caption*{\scriptsize Input image}
        \end{minipage}
        \hfill
        \begin{minipage}[t]{0.24\textwidth}
            \includegraphics[width=\textwidth]{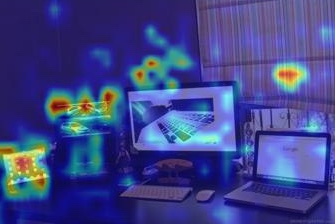}
            \caption*{\scriptsize LLaVA-1.5\\ \textit{There are \textcolor{red}{three} computer monitors in this picture.}}
        \end{minipage}
        \hfill
        \begin{minipage}[t]{0.24\textwidth}
            \includegraphics[width=\textwidth]{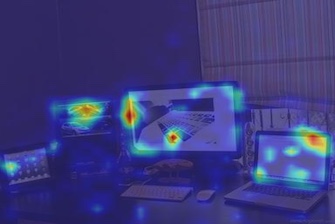}
            \caption*{\scriptsize LLaVA-OV \\ \textit{\textcolor{red}{3}}}
        \end{minipage}
        \hfill
        \begin{minipage}[t]{0.24\textwidth}
            \includegraphics[width=\textwidth]{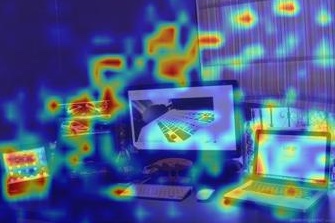}
            \caption*{\scriptsize Cambrian \\ \textit{There are \textcolor{red}{three} computer monitors visible in the picture. The \textcolor{red}{first} is \textcolor{red}{a smaller} monitor to the left, the second is a larger central monitor, and the third is a \textcolor{red}{laptop screen on} the right.}}
        \end{minipage}
        
    \end{subfigure}
    \captionsetup{justification=justified}
    \caption{Qualitative results of the responses of different models and the corresponding focus regions on the image, including attribute (the first and second row) and counting questions (the third row).}
    \label{sup:fig_examples1}
\end{figure*}

\begin{figure*}[htbp]
    \captionsetup{justification=centering}
    \centering
    \captionsetup{font=small}
    \begin{subfigure}{0.49\textwidth}
        \centering
        \caption*{Question: Who is wearing black pants in the image?}
        \begin{minipage}[t]{0.24\textwidth}
            \includegraphics[width=\textwidth]{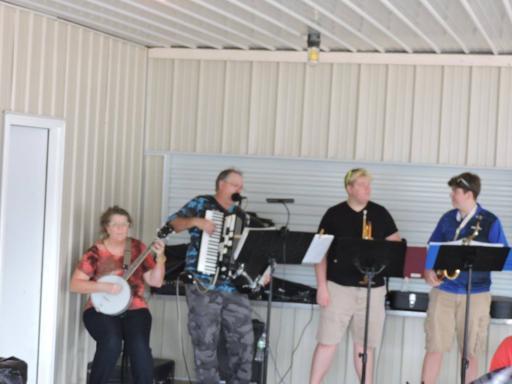}
            \caption*{\scriptsize Input image}
        \end{minipage}
        \begin{minipage}[t]{0.24\textwidth}
            \includegraphics[width=\textwidth]{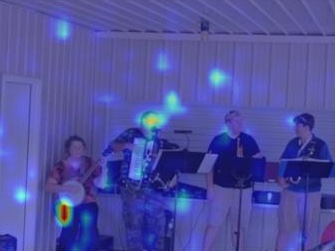}
            \caption*{\scriptsize LLaVA-1.5\\ \textit{A \textcolor{red}{man} is wearing black pants in the image.}}
        \end{minipage}
        \begin{minipage}[t]{0.24\textwidth}
            \includegraphics[width=\textwidth]{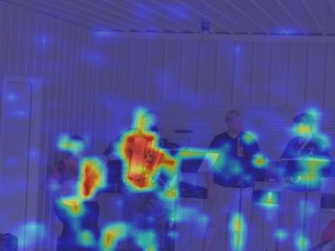}
            \caption*{\scriptsize LLaVA-OV \\ \textit{The man \textcolor{red}{playing the accordion}}}
        \end{minipage}
        \begin{minipage}[t]{0.24\textwidth}
            \includegraphics[width=\textwidth]{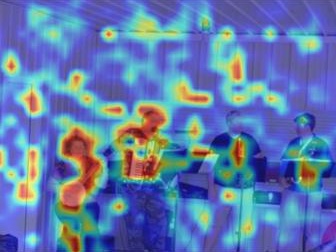}
            \caption*{\scriptsize Cambrian \\ \textit{The individual wearing black pants in the image is the person \textcolor{red}{playing} the \textcolor{red}{accordion}.}}
        \end{minipage}

    \end{subfigure}
    \hfill
    \begin{subfigure}{0.49\textwidth}
        \centering
        \caption*{Question: Which country is highlighted?}
        \begin{minipage}[t]{0.24\textwidth}
            \includegraphics[width=\textwidth]{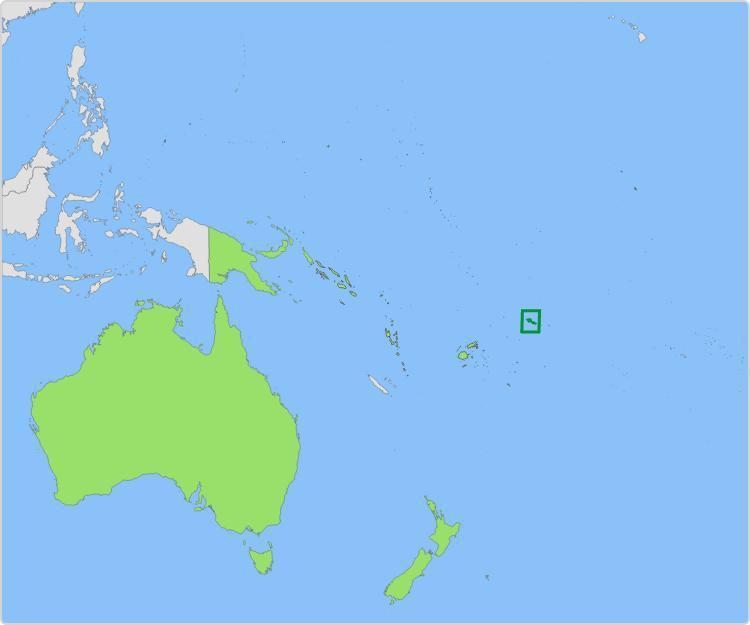}
            \caption*{\scriptsize Input image}
        \end{minipage}
        \hfill
        \begin{minipage}[t]{0.24\textwidth}
            \includegraphics[width=\textwidth]{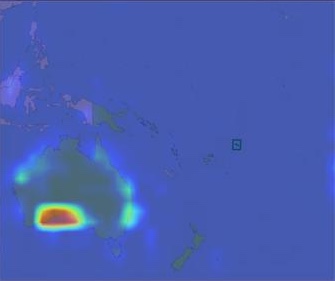}
            \caption*{\scriptsize LLaVA-1.5\\ \textit{The country highlighted in the image is \textcolor{red}{Australia}.}}
        \end{minipage}
        \hfill
        \begin{minipage}[t]{0.24\textwidth}
            \includegraphics[width=\textwidth]{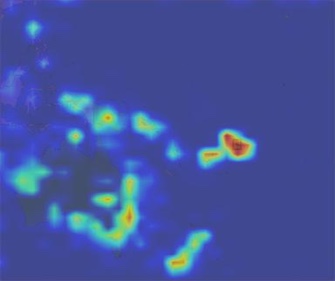}
            \caption*{\scriptsize LLaVA-OV \\ \textit{The \textcolor{red}{highlighted} country is \textcolor{red}{Solomon} Islands.}}
        \end{minipage}
        \hfill
        \begin{minipage}[t]{0.24\textwidth}
            \includegraphics[width=\textwidth]{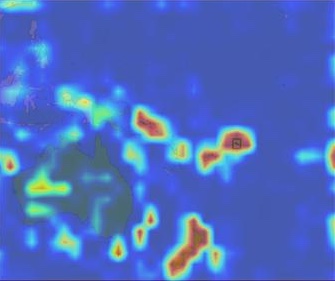}
            \caption*{\scriptsize Cambrian \\ \textit{The country highlighted in the image is \textcolor{red}{N}auru. It is a small island nation located in the South Pacific Ocean, known for its unique political structure and its role in international diplomacy.}}
        \end{minipage}
    \end{subfigure}
    \begin{subfigure}{0.49\textwidth}
        \centering
        \caption*{Question: What is at the top of the food web?}
        \begin{minipage}[t]{0.24\textwidth}
            \includegraphics[width=\textwidth]{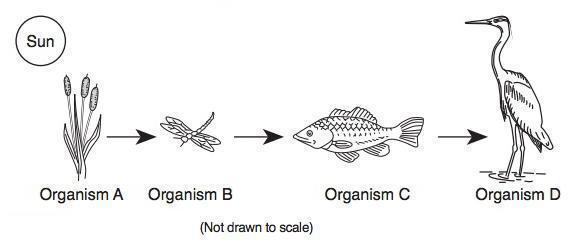}
            \caption*{\scriptsize Input image}
        \end{minipage}
        \begin{minipage}[t]{0.24\textwidth}
            \includegraphics[width=\textwidth]{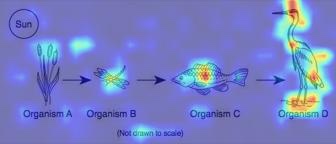}
            \caption*{\scriptsize LLaVA-1.5\\ \textit{At the top of the food web, there is a \textcolor{red}{large bird}, which is likely a \textcolor{red}{her}on or a similar species.}}
        \end{minipage}
        \begin{minipage}[t]{0.24\textwidth}
            \includegraphics[width=\textwidth]{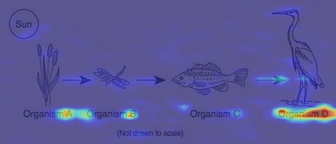}
            \caption*{\scriptsize LLaVA-OV \\ \textit{Organ\textcolor{red}{ism D}}}
        \end{minipage}
        \begin{minipage}[t]{0.24\textwidth}
            \includegraphics[width=\textwidth]{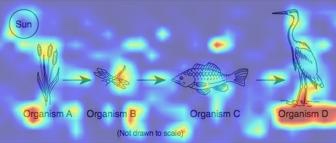}
            \caption*{\scriptsize Cambrian \\ \textit{In the depicted food web, the top predator is the \textcolor{red}{bird}, which is \textcolor{red}{labeled} as \textcolor{red}{Organ}ism \textcolor{red}{D}. This bird is shown as the \textcolor{red}{final} consumer in this particular ecosystem, preying on \textcolor{red}{Organ}ism C, the \textcolor{red}{fish}.}}
        \end{minipage}
    \end{subfigure}
    \hfill
    \begin{subfigure}{0.49\textwidth}
        \centering
        \caption*{Question: Which label marks the egg phase?}
        \begin{minipage}[t]{0.24\textwidth}
            \includegraphics[width=\textwidth]{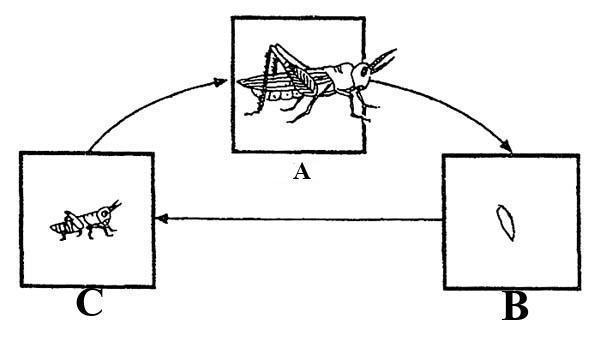}
            \caption*{\scriptsize Input image}
        \end{minipage}
        \hfill
        \begin{minipage}[t]{0.24\textwidth}
            \includegraphics[width=\textwidth]{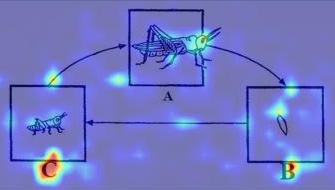}
            \caption*{\scriptsize LLaVA-1.5\\ \textit{The \textcolor{red}{l}abel that marks the egg phase is labeled "\textcolor{red}{C}."}}
        \end{minipage}
        \hfill
        \begin{minipage}[t]{0.24\textwidth}
            \includegraphics[width=\textwidth]{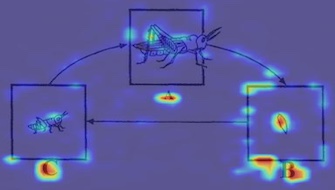}
            \caption*{\scriptsize LLaVA-OV \\ \textit{The egg phase is marked by label \textcolor{red}{B}.}}
        \end{minipage}
        \hfill
        \begin{minipage}[t]{0.24\textwidth}
            \includegraphics[width=\textwidth]{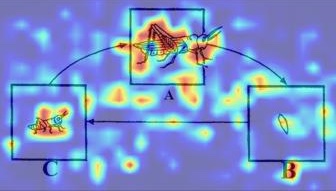}
            \caption*{\scriptsize Cambrian \\ \textit{In the \textcolor{red}{diagram}, the label '\textcolor{red}{B}' \textcolor{red}{corresponds} to the egg phase of the grasshopper's life cycle. This is the stage where the \textcolor{red}{grass}hopper is in its embryonic form, protected within the egg, before it hatches into a nymph.}}
        \end{minipage}
    \end{subfigure}
    \begin{subfigure}[t]{0.49\textwidth}
        \centering
        \caption*{Question: What is the position of the woman in relation to the man in the image?}
        \begin{minipage}[t]{0.24\textwidth}
            \includegraphics[width=\textwidth]{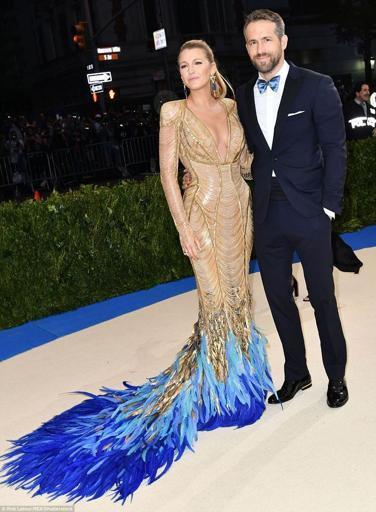}
            \caption*{\scriptsize Input image}
        \end{minipage}
        \hfill
        \begin{minipage}[t]{0.24\textwidth}
            \includegraphics[width=\textwidth]{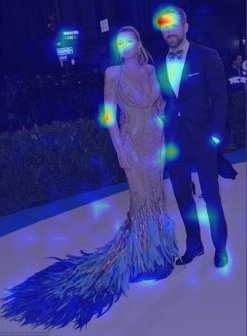}
            \caption*{\scriptsize LLaVA-1.5\\ \textit{The woman is standing \textcolor{red}{next} to the man in the image.}}
        \end{minipage}
        \hfill
        \begin{minipage}[t]{0.24\textwidth}
            \includegraphics[width=\textwidth]{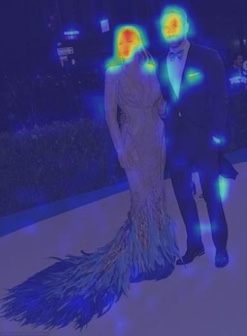}
            \caption*{\scriptsize LLaVA-OV \\ \textit{The woman is standing \textcolor{red}{to} the left of the man in the image.}}
        \end{minipage}
        \hfill
        \begin{minipage}[t]{0.24\textwidth}
            \includegraphics[width=\textwidth]{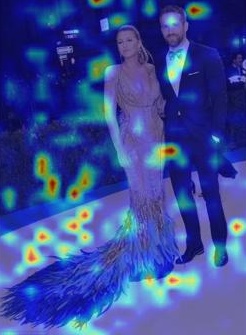}
            \caption*{\scriptsize Cambrian \\ \textit{The woman is standing \textcolor{red}{next} to the man in the image.}}
        \end{minipage}
    \end{subfigure}
    \hfill
    \begin{subfigure}[t]{0.49\textwidth}
        \centering
        \caption*{Question: Which object is closer to the bowl (highlighted by a red box), the microwave (highlighted by a blue box) or the bin (highlighted by a green box)?}
        \begin{minipage}[t]{0.24\textwidth}
            \includegraphics[width=\textwidth]{179_0_image.jpg}
            \caption*{\scriptsize Input image}
        \end{minipage}
        \begin{minipage}[t]{0.24\textwidth}
            \includegraphics[width=\textwidth]{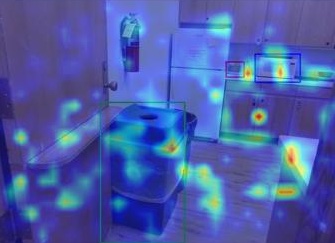}
            \caption*{\scriptsize LLaVA-1.5\\ \textit{In the image, the \textcolor{red}{bin} (highlighted by a green box) is closer to the bowl (highlighted by a red box) than the microwave (highlighted by a blue box). The bin is positioned next to the bowl, while the microwave is located further away in the \textcolor{red}{kitchen}.}}
        \end{minipage}
        \begin{minipage}[t]{0.24\textwidth}
            \includegraphics[width=\textwidth]{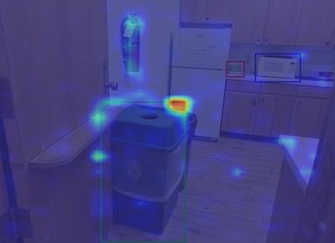}
            \caption*{\scriptsize LLaVA-OV \\ \textit{The \textcolor{red}{bin} (highlighted by a green box) is closer to the bowl (highlighted by a red box).}}
        \end{minipage}
        \begin{minipage}[t]{0.24\textwidth}
            \includegraphics[width=\textwidth]{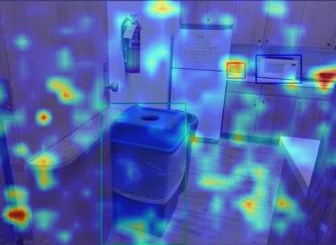}
            \caption*{\scriptsize Cambrian \\ \textit{The microwave, highlighted by the blue box, is closer to the bowl than the bin. The bowl is positioned \textcolor{red}{on} the countertop, and the microwave is directly adjacent to it, while the bin is further away, placed \textcolor{red}{on} the floor.}}
        \end{minipage}

    \end{subfigure}
    \captionsetup{justification=justified}
    \caption{Qualitative results of the responses of different models and the corresponding focus regions on the image, including global (the first row), reasoning (the second row) and spatial (the third row) questions.}
    \label{sup:fig_examples2}
\end{figure*}

\begin{figure*}[tbp]
    \captionsetup{justification=centering}
    \centering
    \captionsetup{font=small}
        \begin{subfigure}{0.49\textwidth}
        \centering
        \caption*{Question: \textit{Does the person in the picture have something in their hand?}}
        \begin{minipage}[t]{0.24\textwidth}
            \includegraphics[width=\textwidth]{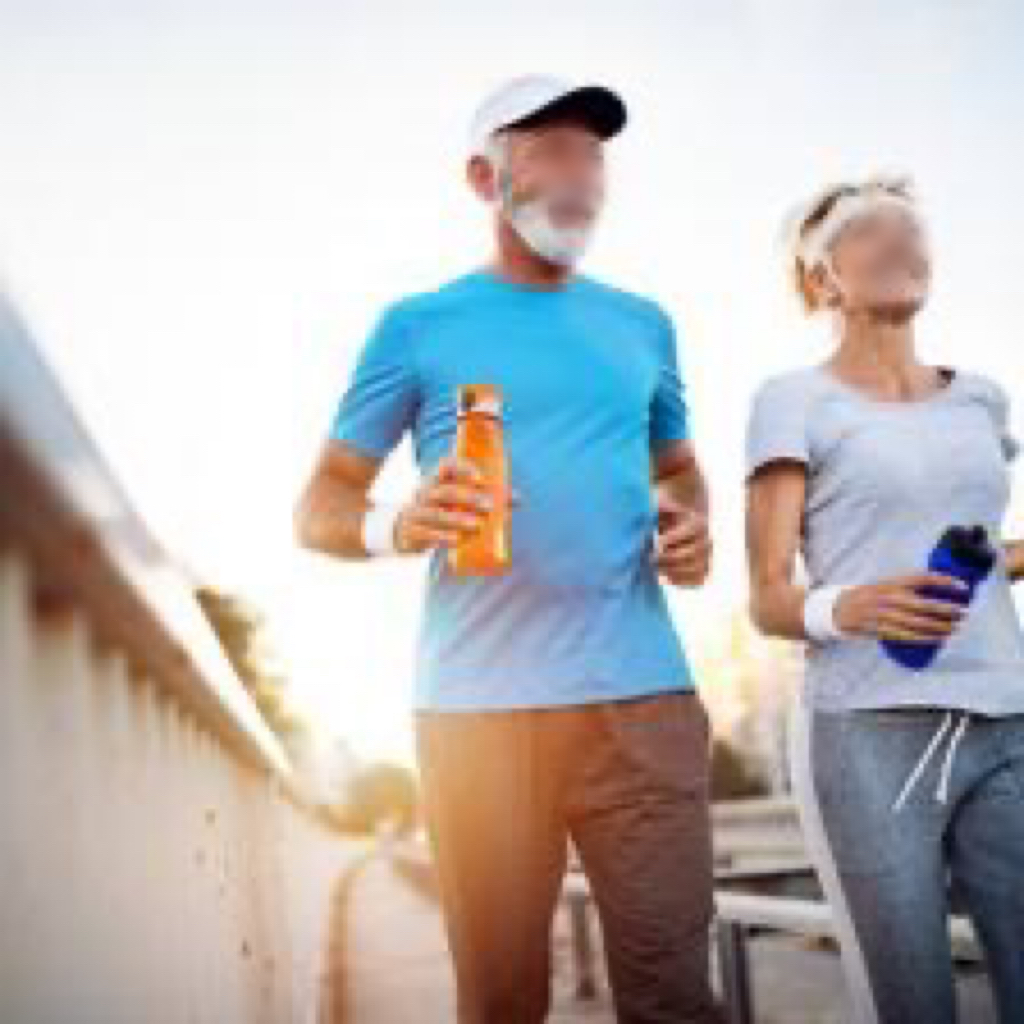}
            \caption*{\scriptsize Input image}
        \end{minipage}
        \begin{minipage}[t]{0.24\textwidth}
            \includegraphics[width=\textwidth]{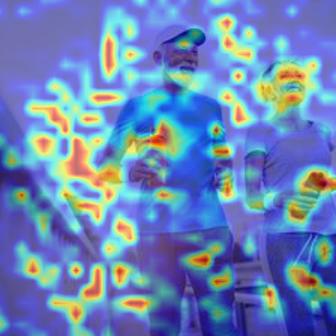}
            \caption*{\scriptsize LLaVA-1.5\\ \textit{Yes, the person in the picture is holding a \textcolor{red}{water} bottle in their hand.}}
        \end{minipage}
        \begin{minipage}[t]{0.24\textwidth}
            \includegraphics[width=\textwidth]{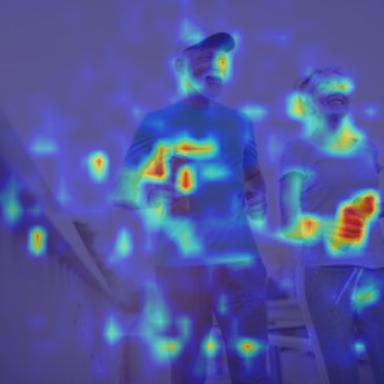}
            \caption*{\scriptsize LLaVA-OV \\ \textit{Yes, the person in the picture is holding a \textcolor{red}{water} bottle.}}
        \end{minipage}
        \begin{minipage}[t]{0.24\textwidth}
            \includegraphics[width=\textwidth]{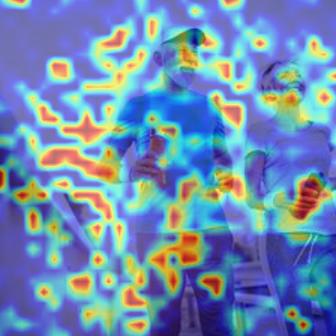}
            \caption*{\scriptsize Cambrian \\ \textit{Yes, the person in the picture is holding a \textcolor{red}{blue water} bottle in their right hand.}}
        \end{minipage}
        \end{subfigure}
        \begin{subfigure}{0.49\textwidth}
        \centering
        \caption*{Question: \textit{What are all the letter in the image?}}
        \begin{minipage}[t]{0.24\textwidth}
            \includegraphics[width=\textwidth]{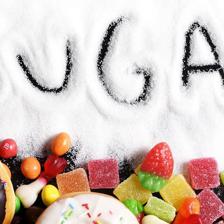}
            \caption*{\scriptsize Input image}
        \end{minipage}
        \begin{minipage}[t]{0.24\textwidth}
            \includegraphics[width=\textwidth]{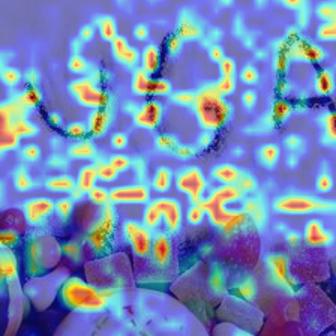}
            \caption*{\scriptsize LLaVA-1.5\\ \textit{The letters in the image are "\textcolor{red}{Sug}ar."}}
        \end{minipage}
        \begin{minipage}[t]{0.24\textwidth}
            \includegraphics[width=\textwidth]{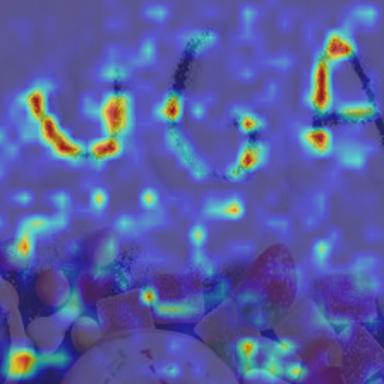}
            \caption*{\scriptsize LLaVA-OV \\ \textit{The \textcolor{red}{letters} in the image are "\textcolor{red}{UG}A".}}
        \end{minipage}
        \begin{minipage}[t]{0.24\textwidth}
            \includegraphics[width=\textwidth]{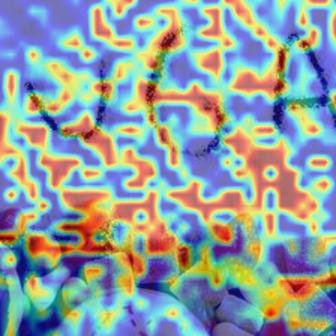}
            \caption*{\scriptsize Cambrian \\ \textit{The image \textcolor{red}{contains} the letters '\textcolor{red}{SUG}AR' \textcolor{red}{spelled} in a stylized manner using what appears to be \textcolor{red}{powdered} sugar.}}
        \end{minipage}
        \end{subfigure}
        \begin{subfigure}{0.49\textwidth}
        \centering
    \caption*{Question: \textit{Is there any accessory on the wrists?}}
        \begin{minipage}[t]{0.24\textwidth}
            \includegraphics[width=\textwidth]{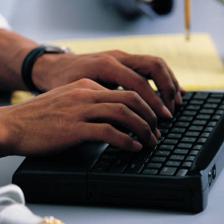}
            \caption*{\scriptsize Input image}
        \end{minipage}
        \begin{minipage}[t]{0.24\textwidth}
            \includegraphics[width=\textwidth]{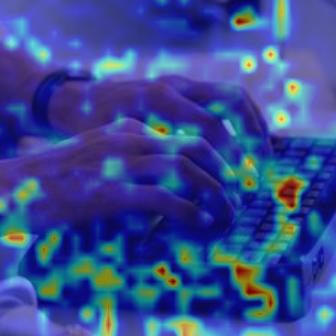}
            \caption*{\scriptsize LLaVA-1.5\\ \textit{Yes, there is a watch on the wrists of the person \textcolor{red}{using} the computer.}}
        \end{minipage}
        \begin{minipage}[t]{0.24\textwidth}
            \includegraphics[width=\textwidth]{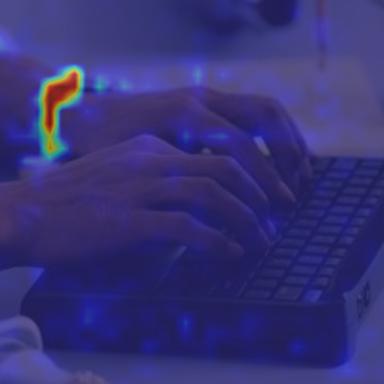}
            \caption*{\scriptsize LLaVA-OV \\ \textit{Yes, there is a \textcolor{red}{black} wristband on the left wrist.}}
        \end{minipage}
        \begin{minipage}[t]{0.24\textwidth}
            \includegraphics[width=\textwidth]{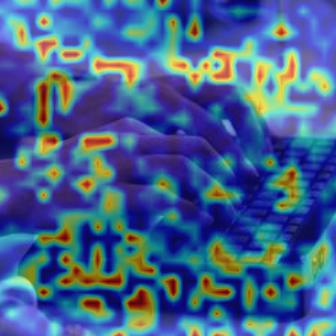}
            \caption*{\scriptsize Cambrian \\ \textit{Yes, the person is wearing a \textcolor{red}{black} wristwatch on their left wrist.}}
        \end{minipage}
    \end{subfigure}
        \begin{subfigure}{0.49\textwidth}
        \centering
        \caption*{Question: \textit{How many trucks are in the image?}}
        \begin{minipage}[t]{0.24\textwidth}
            \includegraphics[width=\textwidth]{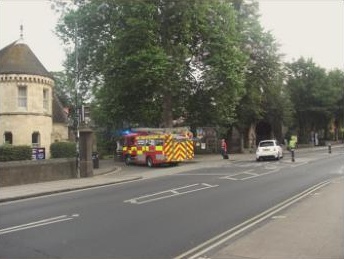}
            \caption*{\scriptsize Input image}
        \end{minipage}
        \begin{minipage}[t]{0.24\textwidth}
            \includegraphics[width=\textwidth]{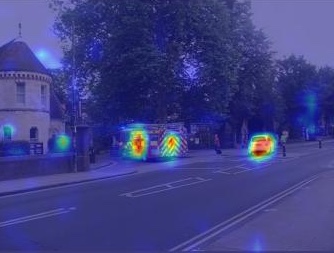}
            \caption*{\scriptsize LLaVA-1.5\\ \textit{There are \textcolor{red}{two} trucks in the image.}}
        \end{minipage}
        \begin{minipage}[t]{0.24\textwidth}
            \includegraphics[width=\textwidth]{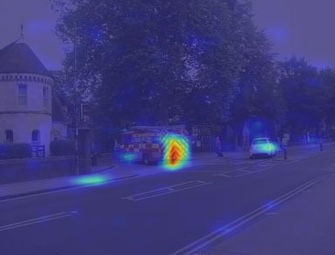}
            \caption*{\scriptsize LLaVA-OV \\ \textit{\textcolor{red}{One}}}
        \end{minipage}
        \begin{minipage}[t]{0.24\textwidth}
            \includegraphics[width=\textwidth]{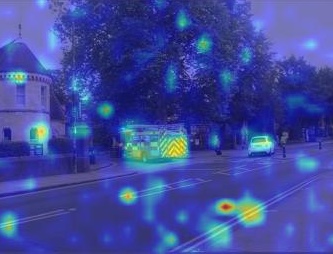}
            \caption*{\scriptsize Cambrian \\ \textit{There is \textcolor{red}{one} truck in the image.}}
        \end{minipage}
        \end{subfigure}
    \caption{Qualitative results of the responses of different models and the corresponding focus regions on the image.}
\label{sup:fig_examples3}
\end{figure*}

In Figure~\ref{sup:fig_examples1},~\ref{sup:fig_examples2} and ~\ref{sup:fig_examples3} we show more qualitative results, comparing the responses and focus regions of LLaVA-1.5, LLaVA-OV and Cambrian. We mainly categorize the questions into spatial, attribute, counting, global and reasoning questions according to their knowledge types. In most cases, the responses of the models well align with the focus regions, especially for some specially contents involved in the responses. For example, in the first example of Figure~\ref{sup:fig_examples2}, the responses of LLaVA-OV and Cambrian mention additional detail of the ``accordion'', which is also highlighted in their heatmaps. Regarding counting questions, for all compared models it is difficult to precisely locate the target objects, hence may lead to wrong answers, as shown in the third row of Figure~\ref{sup:fig_examples1}. For spatial and reasoning questions, we observe that in some cases even though the models focus on the most relevant regions, they still may fail to give the correct answer, since these questions may have higher requirements of spatial reasoning and scientific knowledge.

\section{Limitations}
\label{sec:app_limitation}
In this paper we propose a visually relevant token selection method and extend existing interpretation methods to support open-ended responses of LVLMs with several technical improvements. It illustrates the model generation by deriving a heatmap of the focus region on the image, and can be applied to various model structures with multi-encoder and multi-resolution. 
In this section, we discuss about the potential limitations of our proposed method. 
Since the objective function aims to minimize the deletion score (\emph{i.e.}, replacing pixels in the original image with those in the baseline image) and maximize the insertion score (\emph{i.e.}, replacing pixels in the baseline image with those in the original image), the optimization may not be effective when the output scores given the original image and baseline image do not have significant difference. This is also natural that we can not get the focus region if the responses of the model are not highly related to the input image. In such cases, other interpretation methods can be used as complementary.

\newpage
\clearpage
\newpage
\clearpage

